\RequirePackage{fix-cm}
\documentclass[twocolumn]{svjour3}          
\smartqed  
\usepackage{graphicx}
\usepackage{cite}
\usepackage{breakcites}
\usepackage{times}
\usepackage{epsfig}
\usepackage{epstopdf}
\usepackage{amsmath}
\usepackage{amssymb}
\usepackage{caption}
\usepackage{tabularx}
\usepackage[table]{xcolor}
 \usepackage[normalem]{ulem}
\usepackage{algorithm}
\usepackage[noend]{algpseudocode}
\usepackage[misc]{ifsym}
\usepackage[section]{placeins}
\usepackage{subfig}
\usepackage[pagebackref=true,breaklinks=true,colorlinks,bookmarks=false,citecolor=green]{hyperref}
\usepackage{bm}
\usepackage{multirow} 
\usepackage{booktabs}
\usepackage{comment}
\usepackage{natbib}
\usepackage[table]{xcolor}
\definecolor{mygray}{gray}{.92}

\makeatletter
\newcommand{\thickhline}{%
    \noalign {\ifnum 0=`}\fi \hrule height 1pt
    \futurelet \reserved@a \@xhline
}

\makeatletter
\def\BState{\State\hskip-\ALG@thistlm}
\makeatother
\definecolor{myGreen}{HTML}{33FF00}
\definecolor{myRed}{HTML}{FF3030}
\definecolor{myGrey}{HTML}{AA5555}
\definecolor{myWhite}{HTML}{FFFFFF}
\definecolor{maroon}{cmyk}{0,0.87,0.68,0.32}
\definecolor{petr}{HTML}{5555FF}
\definecolor{josef}{HTML}{FF3030}

\newcommand*{\newcite}[1]{~\citep{#1}}

\journalname{}
\begin{document}

\title{Generalizable Person Search on Open-world User-Generated Video Content}


\author{Junjie Li$^{\dagger}$, Guanshuo Wang$^{\dagger}$, Yichao Yan$^{(\textrm{\Letter})}$, Fufu Yu, Qiong Jia, Jie Qin, Shouhong Ding, and Xiaokang Yang
}


\institute{
J. Li, Y. Yan and X. Yang \at 
are with MoE Key Lab of Artificial Intelligence, AI Institute, Shanghai Jiao Tong University, Shanghai, China. \\
    \email{\{junjieli00, yanyichao, xkyang\}@sjtu.edu.cn}
\and
G. Wang, F. Yu, Q. Jia, and S, Ding \at 
are with Tencent Youtu Lab, Shanghai, China. \\
    \email{\{mediswang,fufuyu,boajia,ericshding\}@tencent.com}
\and
J. Qin \at
is with College of Computer Science and Technology, Nanjing University of Aeronautics and Astronautics, Nanjing, China. \\
    \email{qinjiebuaa@gmail.com}
\and
${}^{(\textrm{\Letter})}$ corresponding author: Y. Yan \at
    \email{yanyichao@sjtu.edu.cn}
\and
${}^{\dagger}$ means equal contribution
}
\vspace{-5mm}
\date{Received: date / Accepted: date}

\maketitle


\begin{abstract}
Person search is a challenging task that involves detecting and retrieving individuals from a large set of un-cropped scene images. Existing person search applications are mostly trained and deployed in the same-origin scenarios. However, collecting and annotating training samples for each scene is often difficult due to the limitation of resources and the labor cost. Moreover, large-scale intra-domain data for training are generally not legally available for common developers, due to the regulation of privacy and public security. Leveraging easily accessible large-scale User Generated Video Contents (\emph{i.e.} UGC videos) to train person search models can fit the open-world distribution, but still suffering a performance gap from the domain difference to surveillance scenes. In this work, we explore enhancing the out-of-domain generalization capabilities of person search models, and propose a generalizable framework on both feature-level and data-level generalization to facilitate downstream tasks in arbitrary scenarios. 
Specifically, we focus on learning domain-invariant representations for both detection and ReID by introducing a multi-task prototype-based domain-specific batch normalization, and a channel-wise ID-relevant feature decorrelation strategy. We also identify and address typical sources of noise in open-world training frames, including inaccurate bounding boxes, the omission of identity labels, and the absence of cross-camera data. Our framework achieves promising performance on two challenging person search benchmarks without using any human annotation or samples from the target domain.

\keywords{Person search \and domain generalization \and open-world data.}
\end{abstract}

\section{Introduction}
Person search is a challenging task that involves simultaneous detection and retrieval of the query individual from an extensive corpus of un-cropped scene images. In supervised person search, the availability of large-scale training datasets, which pertain to the same domain as the target domain, is indispensable for optimal performance due to lacking generalization ability in open-world scenarios. However, the collection of target datasets in different scenarios is frequently impeded by significant resource constraints. Additionally, even when training data from the closely aligned domain is obtained, the manual annotations of precise and high-fidelity ground truth bounding boxes and person identities is a strenuous and labor-intensive task. These constraints impose limitations on the quantity and quality of the available data for model training.

A straightforward solution to the aforementioned challenges is the utilization of large-scale surveillance data for training. However, the collection and utilization of such data can introduce intricate risks related to legal compliance, privacy rights, and security\footnote{\url{https://www.ft.com/content/cf19b956-60a2-11e9-b285-3acd5d43599e}}. In contrast, open-world videos available on public platforms offer greater accessibility and much lower associated risks. A question is naturally raised: \textit{Can we leverage readily available open-world data to train person search models?}
Recent advancements in person re-identification (ReID) pre-training~\newcite{DBLP:conf/cvpr/Fu0BYY0L021,DBLP:conf/cvpr/Fu0YBY0LW022} field have utilized diverse human crops from User-Generated Video Contents (UGC) platforms like YouTube and TikTok for conducting large-scale self-supervised pre-training. Inspired by previous works, we argue that the person search task, characterized by its open-set learning nature, would significantly benefit from the incorporation of UGC video data.
In this paper, we present the first attempt at training person search models using large-scale UGC videos without man-crafted curation. Nevertheless, it is important to acknowledge the existence of a domain gap between UGC videos and surveillance data, including variations in environmental conditions, camera views, and image resolution. These disparities hinder the direct generalization of such models to surveillance scenarios.
To narrow the domain gaps in person search, the unsupervised domain adaptation methods~\newcite{DBLP:conf/eccv/LiYWYJD22} have been proposed. This line of work involves training on a combination of labeled source domain data and unlabeled target domain data. However, they are not suitable for our setting for two key reasons. (1) It is important to note that domain adaptive methods necessitate additional retraining with unlabeled data from the target domain when encountering new scenes. This characteristic imposes a laborious process of data collection and model retraining for each specific deployed scene. (2) Existing domain adaptive methods presume the reliability and high quality of training sources, which are typically human-annotated surveillance data. However, this assumption does not hold when we train models on UGC videos. In a word, these approaches primarily focus on adapting to the target domain rather than improving generalization to arbitrary scenarios.

\begin{figure}[t]
\setlength{\abovecaptionskip}{8pt}
\centering
\includegraphics[width=\linewidth]{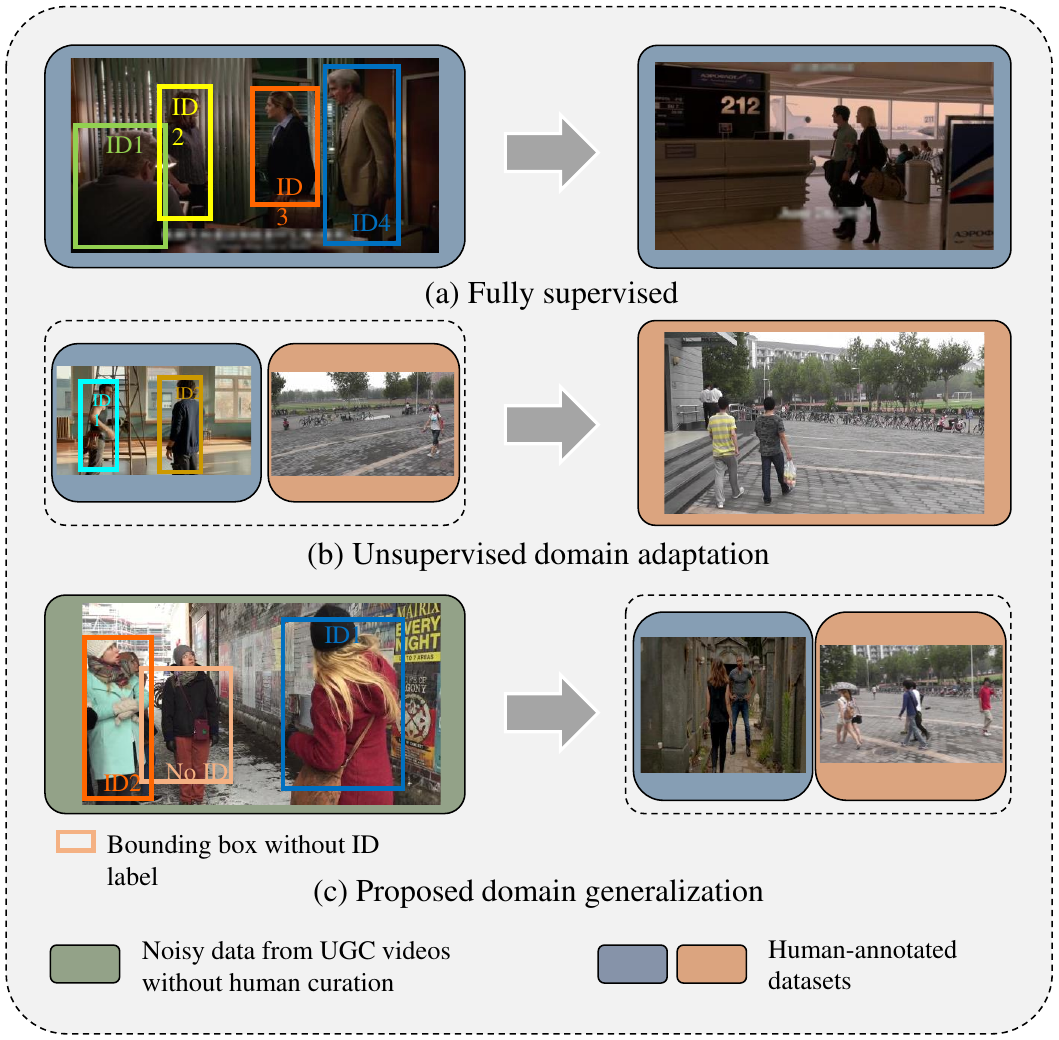}
\caption{Comparing our proposed generalizable person search setting to existing settings. The left and right column represents the training and test set, respectively.}
\label{fig:setting}
\end{figure}

\begin{figure*}[t]
    \centering
    \includegraphics[width=\linewidth]{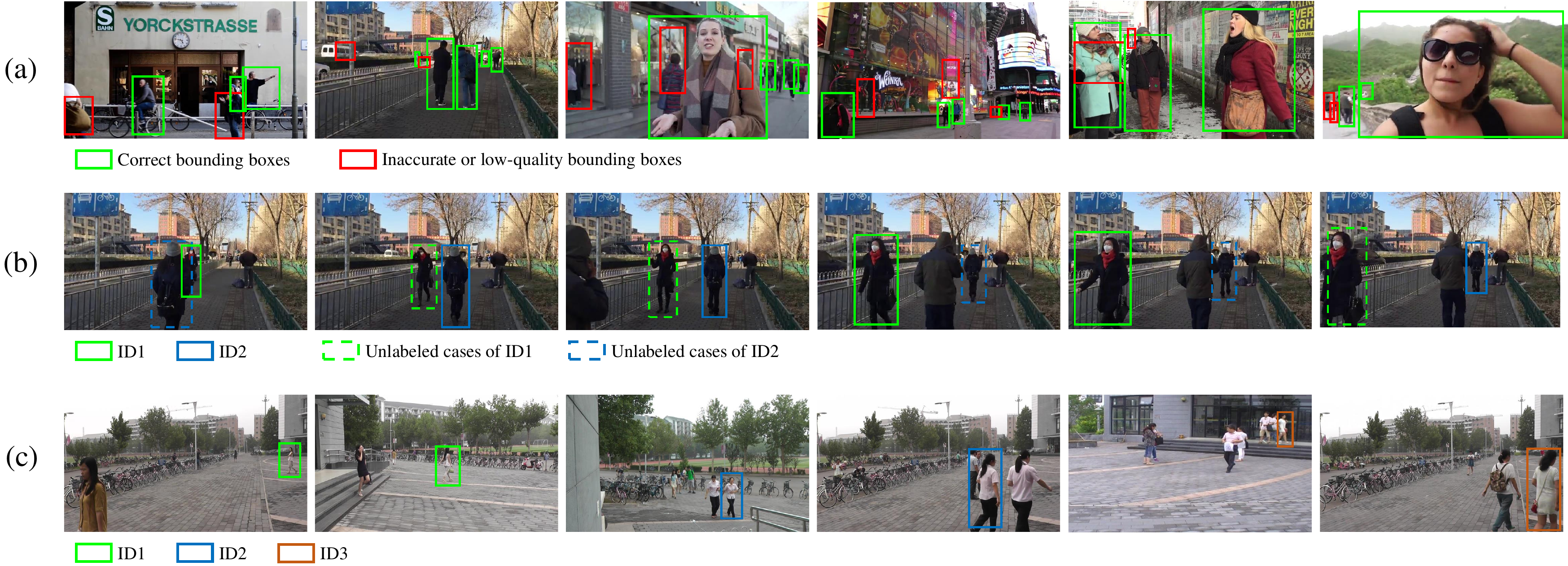}
    \caption{Visualization of typical domain characteristics in UGC and human annotated surveillance videos. (a): The inaccurate bounding boxes annotation, which are marked by red boxes. (b): Identity label omission in neighbouring frames, and the omitted instances are indicated with dashed line. (c): Common cross-camera cases in surveillance data, (\emph{e.g.} PRW dataset), which are almost non-existent in the training data from UGC videos. For simplicity, we do not mark the bounding boxes without ID label in figure (b) and (c).}
    \label{fig:noisy}
\end{figure*}

Alternatively, as shown in Figure~\ref{fig:setting}, we propose the domain generalizable person search setting to apply open-world frames from large-scale UGC videos with proper generalization strategy.
Our proposed setting faces challenges from two aspects: (1) the difficulty of transferring to arbitrary unseen target domains for both the detection and ReID sub-tasks; and (2) the inevitable noise in the automatically generated labels of UGC videos, given the absence of human annotations and curation.

To mitigate the aforementioned issues, we explore both feature-level and data-level generalization solutions. At the feature level, we focus on learning domain-invariant representations for both detection and ReID. To achieve this, we first propose a novel strategy to identify the channel-wise ID-relevant dimensions. On such basis, we further investigate kernel measures of independence~\newcite{DBLP:conf/nips/GrettonFTSSS07,DBLP:journals/jmlr/SongSGBB12}, and decorrelate the statistical dependence between separated ID-relevant and irrelevant representations.
Additionally, in order to fully leverage the diversity of training sources and improve the generalization of both sub-tasks, we divide the UGC training sources based on their inherent metadata. A novel multi-task prototype-based domain-specific batch normalization method is introduced to facilitate invariant representation learning.
At the data level, we have identified three common sources of noise in UGC data, as illustrated in Figure~\ref{fig:noisy}, \emph{i.e.}, inaccurate bounding boxes, omission of identities labels, and scarcity of cross-camera cases. Such data noises can potentially result in optimization dilemma and external domain bias.
To address these issues, we develop targeted methods for each source of noise. Firstly, we address inaccurate bounding boxes by refining regression through the application of soft constraints on low-confidence bounding boxes, effectively reducing their perturbation. To handle the label omission in neighbouring frames, we generate pseudo labels for unlabeled instances by formulating the assignment as a bipartite matching problem. Lastly, to address the scarcity of cross-camera instances, we enhance inter-frame cases and distinguish intra-frame samples to fully unleash the potential of available data. These operations work seamlessly to improve the generalization of both the detection and ReID sub-tasks on the open-world training sources.

Our main contributions are summarized as follows:
\begin{itemize}
    \setlength{\itemsep}{2pt}
	\setlength{\parsep}{-3pt}
	\setlength{\parskip}{-0pt}
	\setlength{\leftmargin}{-15pt}
    \item To the best of our knowledge, this is the first attempt to train a person search model using un-cropped frames from large-scale open-world videos without any human annotation. On such basis, we first introduce a novel and practical domain generalization setting for the person search task.

    \item We present feature-level invariant representation learning and data-level generalization on noisy sources to alleviate the domain gaps for both sub-tasks and improve the generalization capability in the presence of noisy training data.

    \item Without any human-crafted annotation and samples from the target domain, our framework achieves promising performance on two target person search benchmarks and largely outperforms the direct transfer baseline. Our work validates the feasibility of training a person search model with open-world data and deploying it to arbitrary scenes.
    
\end{itemize}

\section{Related Work}
\subsection{Person Search}
Person search has gained significant attention in recent times, particularly with the emergence of large-scale surveillance benchmarks~\newcite{DBLP:conf/cvpr/XiaoLWLW17,DBLP:conf/cvpr/ZhengZSCYT17}. Fully supervised person search models can be broadly classified into two categories: \textbf{(1)} Two-step approaches that combine separately trained pedestrian detection and ReID models. PRW~\newcite{DBLP:conf/cvpr/ZhengZSCYT17} systematically evaluated combinations of independent sub-task models. IGPN~\newcite{DBLP:conf/cvpr/DongZST20} decreased detection proposals with similarity scores. TCTS~\newcite{DBLP:conf/cvpr/WangMCSC20} addressed the inconsistency between detection and ReID sub-tasks. \textbf{(2)} One-step methods that jointly integrate detection and ReID in an end-to-end framework. One-step frameworks generally offer notable advantages in terms of efficiency and convenience compared to two-step methods. OIM~\newcite{DBLP:conf/cvpr/XiaoLWLW17} first introduced a Faster RCNN based pipeline with an 
integrated ReID branch. Most subsequent methods~\newcite{DBLP:conf/cvpr/ZhongWZ20,DBLP:conf/cvpr/MunjalATG19,DBLP:journals/ijcv/ChenZYS21,DBLP:conf/cvpr/YanZNZXY19,DBLP:conf/aaai/ChenZO0S20,DBLP:conf/wacv/JaffeZ23} have followed this line of work. ROI-AlignPS~\newcite{DBLP:journals/ijcv/YanLQZLY23} design a novel anchor-free approach for one-step person search, and SeqNet~\newcite{DBLP:conf/aaai/LiM21} introduce an extra preposed detection module to improve proposal quality for subsequent ReID branch. Recently, some works~\newcite{DBLP:conf/cvpr/CaoPAC0SK22,DBLP:conf/cvpr/YuDLDFHC22,DBLP:journals/corr/abs-2211-04323} further developed transformer-based frameworks and achieved promising performance. However, such fully supervised methods typically require expensive human labeling and show unsatisfactory performance when tested on different scenarios.

To avoid laborious human annotation on each novel scene, researchers have developed two alternative settings: weakly supervised~\newcite{DBLP:conf/iccv/HanSYYGS0W21,DBLP:conf/aaai/YanLLQNLY22,DBLP:conf/bmvc/HanKS21} and domain adaptive~\newcite{DBLP:conf/eccv/LiYWYJD22}. In the weakly supervised person search approach, bounding boxes are used as the only source of supervision, and the pseudo labels generated from clustering are used to supervise ReID sub-task. Conversely, DAPS~\newcite{DBLP:conf/eccv/LiYWYJD22} propose a domain adaptive setting where the model is trained using both labeled source and unlabeled target domains. However, these frameworks have not gotten rid of retraining on each scene. In this work, we take a step further by introducing a more applicable domain generalization scenario. Under this setting, the target domain is unseen during training, and the generalization ability of the model is highly challenged.

\subsection{Domain Generalization}
Most deep learning-based approaches rely on the assumption of independently and identically distributed (i.i.d.) data, which is often violated in open-world applications due to widely prevalent domain shift. To address this issue, the domain generalization (DG)~\newcite{zhou2022domain} technique has been proposed, which aims to learn a robust model using only the source data for training. DG has been extensively studied for various vision tasks, \emph{e.g.} image classification~\newcite{DBLP:conf/nips/BalajiSC18,DBLP:conf/icml/MuandetBS13,DBLP:conf/nips/VolpiNSDMS18,DBLP:conf/aaai/LiYSH18,DBLP:journals/ijcv/YuanMCKWL23}, semantic segmentation~\newcite{DBLP:conf/iccv/YueZZSKG19,DBLP:conf/iccv/VolpiM19,DBLP:conf/nips/ZhongZLS22,DBLP:conf/eccv/ZhaoZZSL22}, and object detection~\newcite{DBLP:journals/corr/abs-2203-14387,DBLP:conf/cvpr/ViditES23}.

Recently, several works~\newcite{DBLP:conf/cvpr/SongYSXH19,DBLP:conf/cvpr/ChoiKJPK21,DBLP:journals/corr/abs-2111-14290,DBLP:conf/eccv/LiaoS20,DBLP:conf/eccv/XuLHS22,DBLP:journals/pami/PuZSL23,DBLP:conf/cvpr/Liao022,dbn} have introduced domain generalization to the ReID task. Existing DG ReID methods can be generally divided into two categories: \textbf{(1)} those that aim to learn domain-invariant embeddings from single/multi-source domains. For example, $M^3L$~\newcite{DBLP:conf/cvpr/ZhaoZYLLLS21} proposed a meta-learning strategy to diversify meta-test features and simulate the target domain, and Jin \emph{et al.}~\newcite{DBLP:conf/cvpr/JinLZ0Z20} decoupled the normalized identity feature into id-relevant and id-irrelevant parts. MMFA-AAE~\newcite{DBLP:journals/tip/LinLK21} presented an adversarial auto-encoder to learn a domain-invariant latent feature from multi-sources. \textbf{(2)} those that incorporate the mixture of experts strategy into DG ReID. Dai \emph{et al.}~\newcite{DBLP:conf/cvpr/DaiLLTD21} utilized an effective voting-based mixture mechanism to dynamically leverage source domains. However, the ReID sub-task in person search considers the context and scene information of uncropped images, and this characteristic distinguishes it from DG ReID methods. In this work, we consider both the detection and ReID sub-tasks together. Our framework decorrelate the identity feature in the channel dimension, apply prototype-based domain-specific batch normalization for both sub-tasks, and propose strategies for data-level generalization problem.

\section{Training Dataset}\label{chapter:training_dataset}
The training dataset employed in this study was constructed upon the large-scale LUPerson-NL (LUP-NL) dataset~\newcite{DBLP:conf/cvpr/Fu0YBY0LW022}, which was originally created from raw YouTube videos for the purpose of ReID Pre-training. Despite being automatically annotated with an off-the-shelf tracking algorithm~\newcite{DBLP:journals/ijcv/ZhangWWZL21}, the labels are known to be noisy and sparse for person search. To address these limitations, we randomly selected a subset of LUP-NL, whose samples come from three different countries across different continents. The subset is re-annotated using YOLOv5~\newcite{DBLP:journals/corr/abs-2104-13634} to improve the suitability for person search task. These efforts result in a multi-source training dataset that includes 211,982 images from 680 raw videos, with a total of 10,141 identities. The names of sampled videos are available in the appendix material. It is important to note that all annotations for the training data are generated without human labeling or checking. As a result, we further discuss how to enhance the generalization ability on such this source in section~\ref{chapter:noisy}.

\section{Methodology}
\subsection{Framework Overview}
\begin{figure*}[t]
    \setlength{\abovecaptionskip}{8pt}
    \centering
    \includegraphics[width=\linewidth]{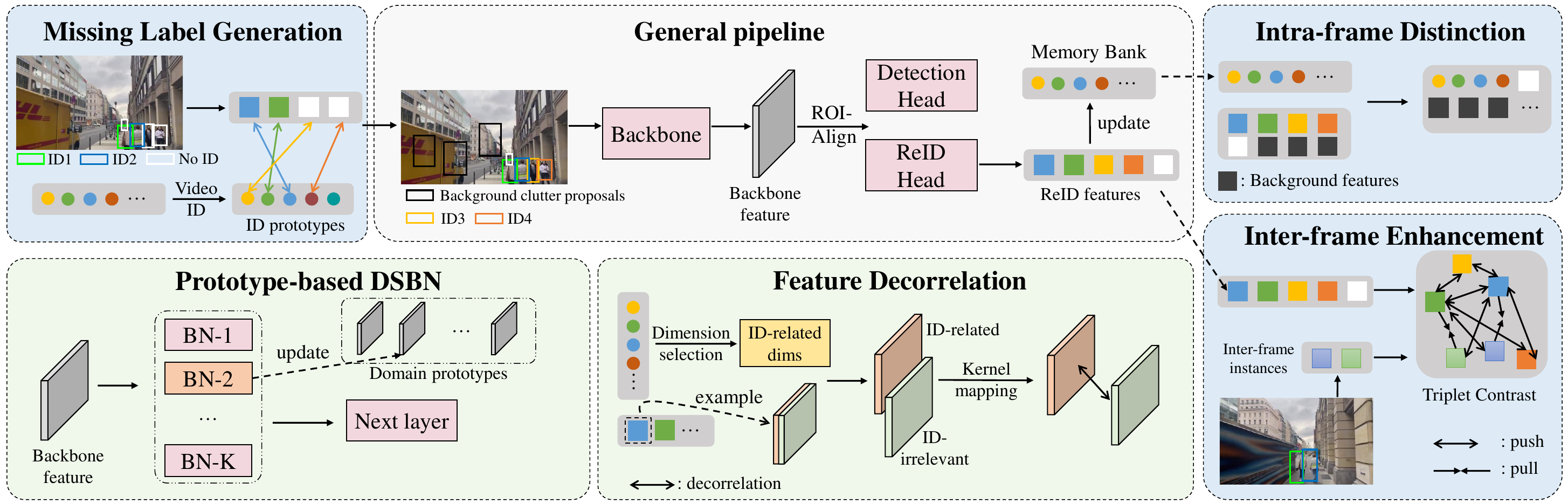}
    \caption{The overall pipeline of our proposed generalizable person search (GPS) framework, and the base pipeline follows SeqNet~\newcite{DBLP:conf/aaai/LiM21}. The components with light green background  are designed for feature-level invariant learning, while those with blue background are utilized for data-level generalization.}
    \label{fig:pipeline}
\end{figure*}
Given an input image $I \subseteq \mathbb{R}^{C \times H \times W}$, where $C, H, W$ denote the number of channels, the height, and the width of the image, respectively. The person search model first needs to predict a set of bounding boxes $B \subseteq \mathbb{R}^{\mathrm{n} \times 4} $, each containing a pedestrian. The identity representations $X \subseteq \mathbb{R}^{\mathrm{n} \times \mathrm{d}} $ of the $\mathrm{n}$ pedestrians are further produced by the ReID head, where $\mathrm{d}$ is fixed to 256 following~\newcite{DBLP:journals/ijcv/ChenZYS21,DBLP:conf/aaai/LiM21}. The person search model is represented as $\mathcal{P} : I \rightarrow (b, x)^\mathrm{n}$, where $b \in B$ and $x \in X$. In our domain generalization scenario, each training sample $I$ is drawn from a source domain with known domain label $d_I \in D$, where $D$ is the set of all training domains. Given the evaluation metrics $\mathcal{M}(I;\mathcal{P};Y)$ of both sub-tasks. The target is to optimize the function $\mathcal{P}$ to maximize the metrics $\mathcal{M}$ on an unseen target domain $d_t \not\in D$.
The overall pipeline of our proposed framework is presented in Figure~\ref{fig:pipeline}.
\subsection{Feature-Level Domain Invariant Learning}
\textbf{ID-relevant Feature Decorrelation}.
We begin by performing a channel-wise division of the identity representation $x$ from the normalized identity features $X \subseteq \mathbb{R}^{\mathrm{n} \times \mathrm{d}}$ into two distinct components: the id-relevant component $x^{id}$ and the domain-specific component $x^{ds}$ based on their statistical dependence. Our optimization objective is to eliminate any statistical dependence between these two divided components. To achieve this objective, we aim to partition the normalized identity features $X$ into $\mathrm{n}$ $\mathrm{d'}$-dimension id-related features $X^{id} \overset{def}{=} \{X^{id}_{1}, X^{id}_{2},..., X^{id}_{\mathrm{n}} \}$ and domain-specific features $X^{ds} \subseteq \mathbb{R}^{\mathrm{n} \times \mathrm{(d-d')}}$, such that any pair of $\{ X^{id}_{:,i}, X^{ds}_{:,j}\}$ exhibits statistical independence. However, the implementation of this approach poses two significant challenges. The first challenge is to effectively separate the id-related dimensions from the domain-specific ones. The second challenge is to eliminate any relevance between the paired variables.

To tackle the first challenge, we apply identity prototypes stored in the memory bank $M \subseteq \mathbb{R}^ {\mathrm{N}^{id} \times \mathrm{d} }$, where $\mathrm{N}^{id}$ represents the number of single-instance classes. To this end, we adopt a classifier $\mathcal{C}$ to predict the identity of each prototype. Evaluation metric $\mathcal{M}^c( M,\mathcal{C})$ is used to obtain the importance values $\alpha = (\alpha_1, \alpha_2, ... , \alpha_\mathrm{d}) \subseteq \mathbb{R}^{\mathrm{d} \times \{0, 1 \}}$. These values determine the relevance of each feature dimension with respect to the identity.
\begin{equation}
\begin{aligned}
    &\operatorname{min}(\Vert \alpha \Vert_1), \\
    \text{ s.t. } \mathbb{E}_{\alpha} [\mathcal{M}^c( &M \cdot \alpha,\mathcal{C})] \geq \mathbb{E} [\mathcal{M}^c( M,\mathcal{C})] - t,
\end{aligned}
\end{equation}
where $t$ is the declining threshold. We update the weights $\alpha$ after each training epoch completes, and treat them as the global decorrelation weights for all samples.

To address the second challenge, we propose to decorrelate both the linear and non-linear connections between $X^{id}$ and $X^{ds}$. To simplify the notation, we represent the randomly selected pair $\{ X^{id}_{:,i}, X^{ds}_{:,j}\}$ as $\{ Z^{id}, Z^{ds} \}$. First, we consider the linear dependencies, which can be measured and eliminated using cosine similarity:
\begin{equation}
\begin{aligned}
    \operatorname{min} \frac{ (Z^{id})^T  Z^{ds}} { \Vert Z^{id} \Vert  \Vert Z^{ds} \Vert},
\end{aligned}
\end{equation}
However, the complexity and non-linearity of person search model makes it challenging to mine the implicit non-linear dependencies. Kernel methods have been widely employed for mapping the low-order data into Reproducing Kernel Hilbert Space (RKHS) for further capturing the feature dependence. Suppose the RKHSs on $Z^{id}$ and $Z^{ds}$ are defined as $\mathcal{H}(Z^{id})$ and $\mathcal{H}(Z^{ds})$, respectively, with kernel methods $f, g$ satisfying:
\begin{equation}
\begin{aligned}
\mathbb{E}&\left[f(Z^{id}, Z^{id})\right]<\infty, \\
\mathbb{E}&\left[g(Z^{ds}, Z^{ds})\right]<\infty, \\
\mathbb{E}_{Z^{id} \sim P}[f(Z^{id})]=&\mathbb{E}_{Z^{id} \sim Q}[f(Z^{id})] \iff P=Q, \\
\mathbb{E}_{Z^{ds} \sim P}[g(Z^{ds})]=&\mathbb{E}_{Z^{ds} \sim Q}[g(Z^{ds})] \iff P=Q,
\end{aligned}
\end{equation}
which guarantees the RKHSs belonging to the the square integrable spaces with probability distributions of $Z^{id}$ and $Z^{ds}$. The independence can be expressed in terms of the cross-covariance operators~\newcite{baker1973joint} $\Sigma$ on RKHS:
\begin{equation}
\begin{aligned}
\left\langle f, \Sigma_{Z^{id} Z^{ds}} g\right\rangle=&\mathbb{E}_{Z^{id} Z^{ds}}[f(Z^{id}) g(Z^{ds})] \\
-&\mathbb{E}_{Z^{id}}[f(Z^{id})] \mathbb{E}_{Z^{ds}}[g(Z^{ds})], 
\end{aligned}
\end{equation}
when in the euclidean space the operator itself can be formulated as:
\begin{equation}
\begin{aligned}
\Sigma_{Z^{id} Z^{ds}} = \mathbb{E}_{Z^{id} Z^{ds}}[ &(f(Z^{id}) - \mathbb{E}_{Z^{id}}f(Z^{id}) )  \\
\otimes & (g(Z^{ds}) - \mathbb{E}_{Z^{ds}}g(Z^{ds}) )].
\end{aligned}
\end{equation}

According to the derivation in~\newcite{DBLP:journals/jmlr/GrettonHSBS05}, $\Sigma_{Z^{id} Z^{ds}}$ is zero if and only if the random variables are independent. Therefore, it is possible to measure the independence~\newcite{DBLP:conf/nips/GrettonFTSSS07,DBLP:journals/jmlr/SongSGBB12} via the cross-covariance operators. Hilbert-Schmidt Independence Criterion (HSIC)~\newcite{DBLP:conf/nips/GrettonFTSSS07} regularizes the square of the Hilbert-Schmidt norm of the operator $\Sigma_{Z^{id} Z^{ds}}$. Since the HS norm equals the Frobenius norm in the euclidean space, the square F-norm of the cross-covariance matrix can be used in the regularization term. Subsequently, we can apply a regularizer to remove the dependency between the identity-related dimensions and domain-specific variables:.
\begin{equation}
\begin{aligned}
L_{cov}=\frac{1}{(\mathrm{n}-1)^2} \operatorname{tr}(F H G H) = \left\|{\Sigma}_{Z^{id} Z^{ds}}\right\|_F^2,
\end{aligned}
\end{equation}
where $H = I-\frac{1}{\mathrm{n}}  {1 1}^{\top}$, and $1$ represents the $k \times 1$ matrix filled with ones. Some shift-invariant kernels, \emph{e.g.}, the Gaussian RBF kernel are utilized in our kernel-based methods. To improve the computational convenience, we test some approximation approaches~\newcite{strobl2019approximate,DBLP:conf/cvpr/Zhang0XZ0S21} via random features~\newcite{DBLP:conf/nips/RahimiR07}.

\textbf{Multi-task Prototype Based DSBN}.
The domain shift problem can be addressed using domain-specific batch normalization (DSBN)~\newcite{DBLP:conf/cvpr/ChangYSKH19}, a technique that normalizes instances from each domain separately through individual BN branches.
However, applying DSBN directly to the generalization scenario is impractical due to the unavailability of the target domain during training. Existing generalizable ReID methods~\newcite{DBLP:conf/cvpr/DaiLLTD21,DBLP:journals/corr/abs-2111-14290} address this issue by separating a meta-test domain from the source domains to simulate the unseen target domain. In this study, we propose a novel prototype-based strategy without maintaining meta-test domain during training, and boost the generalization of both sub-tasks by applying it on both heads.
We formulate DSBN for samples of the $i$-th domain as follows:
\begin{equation}
\operatorname{DSBN}\left(x ; \gamma_i, {\beta_i}\right)_{i}=\gamma_i \cdot \frac{x-{\mu_i}}{\sqrt{{\sigma_i}^2+\epsilon}}+ \beta_i,
\end{equation}
where $\gamma_i$ and $\beta_i$ are parameters corresponding to the $i$-th domain. $\mu_i, \sigma_i$ denote the running meaning and variance with respect to a batch, respectively. 

As shown in Figure~\ref{fig:pipeline}, we maintain a domain prototype for each training domain represented by $C \subseteq \mathbb{R}^{\mathrm{K} \times \mathrm{d}^b}$, where $ \mathrm{d}^b $ is the dimension of features from backbone. The prototypes are updated using a moving average operation with a momentum scalar $m$, as follows:
\begin{equation}
\label{formulation:update}
c \leftarrow m \cdot c+(1-m) x,
\end{equation}
where $c \in C$, and the input $x$ shares the same domain label with $c$. During testing, we estimate the weighting terms $\zeta = (\zeta_1, \zeta_2, ..., \zeta_{\mathrm{K}})$ for a given instance by computing the softmax cosine similarities between the input and $C$. The BN parameters for the input $x \in d_t$ can be written as:
\begin{equation}
\operatorname{DSBN}(x; \zeta, \gamma, \beta)=\sum_{i=1}^K \zeta_i \operatorname{DSBN}(x ; \gamma_i, {\beta_i})_{i}.
\end{equation}
Note that we extend the prototype-based method to the detection sub-task by replacing all the BN layers in the detection head and the ReID head with the DSBN layers.

\subsection{Data-level Generalization on Noisy Sources}\label{chapter:noisy}
Although automatic annotation without manual intervention from media sources is convenient on UGC videos, it can result in noisy training data that deteriorates model training and introduces implicit domain bias. As depicted in Figure~\ref{fig:noisy}, three major challenges arise with noisy sources. Firstly, noisy bounding box annotations can limit the detection performance on the target domain. 
Secondly, omission of identity labels is common in training frames selected from tracklets. Take Figure~\ref{fig:noisy} (b) as an example, when two adjacent frames are generated by different tracklets, a person may only have an identity label in one of the two adjacent frames. While this omission has little impact on ReID with cropped human images, it severely constrains the training of generalizable person search.
Lastly, discriminating cross-camera cases is crucial for generalizable person search models. However, samples with the same identity label typically originate from a continuous segment of the same video, where the variation of background is relatively small. To address the aforementioned challenges, we propose generalization strategies like bounding boxes refinement, generating pseudo-labels, and unleashing the potential of training data by inter-frame Enhancement intra-frame distinction.

\textbf{Bounding Boxes Refinement}.
To enhance the training of the detection sub-task and accommodate the occurrence of noisy annotations, we implemented both hard and soft constraints. More precisely, we merged the initial labels from LUP-NL with the bounding boxes generated by YOLOv5, and subsequently eliminated low-confidence boxes as part of our hard constraint. In addition, we incorporated the annotation confidence as a soft weight for both regression and classification loss functions. These constraints ensure a more accurate and robust detection head by mitigating the impact of noisy data while still enabling the model to learn from them. 

\textbf{Missing Labels Generation}.
Label omission can create conflicts in model optimization, whereby improved detection performance may negatively impact the optimization of the ReID head by identifying missing labeled instances. To address this issue, we propose assigning pseudo identity labels to qualified None-ID instances and eliminate their negative effects.

Suppose that $n$ people with a fixed number of $T$ identities appear in a video source $V$, and let the detected boxes in a frame $v \in V$ be denoted as $B \subseteq \mathbb{R}^{\mathrm{m} \times 4}$. The goal is to assign a pseudo label subset $(t_1, t_2, ...) \subseteq T$ to eligible instances in $B=(b_1, b_2,...)$. Since a person can only appear once in a frame, the optimal assignment can be formulated as a bipartite matching between $B$ and $T$, and the usual solution is to use the Hungarian algorithm~\newcite{kuhn1955hungarian}. We define the weight matrix $H$ for pseudo label generation as follows:
\begin{equation}
\begin{aligned}
H_{i,j} \overset{def}{=}
\begin{cases}
\operatorname{cos}(b_i, t_j)&; \operatorname{ID}(b_i) \in \emptyset, \operatorname{cos}(b_i, t_j) > \psi  \\
1&; \operatorname{ID}(b_i) \in T\\
0 &; \text{elsewise}
\end{cases},
\end{aligned}
\end{equation}
where $\operatorname{ID}()$ is the operation of obtaining the identity label of the given box, $\psi$ is the lower threshold for assigning a pseudo label. The weight matrix shape is then zero padded to $l \times l$, and $l = \operatorname{max}(\operatorname{len}(B), \operatorname{len}(T))$. Once we obtain the weight matrix $H$, pseudo labels are assigned to instances $B$ according to the assignment matrix $A$:
\begin{equation}
\begin{aligned}
&A_{i,j} \overset{def}{=}
\begin{cases}
1&; \operatorname{ID}(b_i) = t_j \\
0 &; \text{elsewise}
\end{cases}, \\
\text{ s.t. } &\forall 0 \leq i \leq l, \sum_{j=0} A_{i, j} \leq 1 ,\\
&\forall 0 \leq j \leq l, \sum_{i=0} A_{i, j} \leq 1 ,
\end{aligned}
\end{equation}
and the assignment follows the objectiveness of maximize $\left\| A \cdot H \right\|_1$, where $\left\| \right\|_1$ denotes the $L_1$ entry-wise norm. In practice, the label generation is scheduled before the start of each epoch after the first one.

\textbf{Inter-frame Enhancement and Intra-frame Distinction}.
The absence of cross-camera training data may introduce an external domain bias that couples the model to specific backgrounds and locations. To overcome this, we propose two methods, namely the `inter-frame enhancement' and `intra-frame distinction'. The former involves simulating cross-camera cases with existing samples, while the latter enhances the distinction between instances with similar backgrounds. When the same identity appears in two frames that are far apart, the background and perspective may differ due to camera view shift. We leverage these instances to simulate cross-camera cases and facilitate matching of identities across cameras. This is achieved by applying a triplet loss for common embedding space of identity features in both anchor frames and inter-frame cases:
\begin{equation}
L_{IE}= \sum_{i} ( m_t+\max _{v \in v_i^{+}} \operatorname{d}\left(x_i, v\right)-\min _{v \in v_i^{-}} \operatorname{d}\left(x_i, v\right) ),
\end{equation}
where $\operatorname{d}$ is a distance function, $m_t$ is the margin parameter, and $v_i^+, v_i^-$ represent the set of positive and negative samples with $x_i$, respectively. However, efficiently sampling inter-frame instances remains a challenge in this context. In this work, we maintain two extra prototypes for each identity, which represent instances appearing in the first half and second half of the frame sequence, respectively. When the anchor $x_i$ comes from a frame belonging to the first half of the sequence, the corresponding momentum updated identity prototype of the second half is employed as the positive sample and vice versa. A further discussion about the reasonableness of such implementation can be found in section~\ref{discussion}.

As for the intra-frame distinction, let us first recall the widely employed OIM loss:
\begin{equation}
L_{oim}=-\log \frac{\exp \left(v_{+}^T x / \tau\right)}{\sum_{j=1}^L \exp \left(v_j^T x / \tau\right)+\sum_{k=1}^Q \exp \left(u_k^T x / \tau\right)},
\end{equation}
where $v$ represents the feature in the identity memory $M$ and $u$ represents the unlabeled instance from the negative sample memory. We can make the observation that the negative sample memory is updated in the backward stage, thus the intra-frame contrast is not optimized explicitly during training. To this end, we introduce intra-frame negative samples and background clutters into the training process. The modified loss is formulated as:
\begin{equation}
\begin{aligned}
L_{ID}=&-\log \frac{\exp \left(v_{+}^T x / \tau\right)} {\sum_{z \in \mathbf{M}}{\exp \left(z^T x / \tau\right)}} \\
\sum_{z \in \mathbf{M}}{\exp \left(z^T x / \tau\right)} = &\sum_{j=1}^L \exp \left(v_j^T x / \tau\right)+\\
&\sum_{k=1}^Q \exp \left(u_k^T x / \tau\right) + \sum_{l=1}^N \exp \left(c_l^T x / \tau\right),
\end{aligned}
\end{equation}
where $c$ denotes the intra-frame negative samples.

\section{Experiment}
\subsection{Experiment Setup}
\textbf{Dataset}.
To simplify the notation, we denote the re-organized training dataset described in Section~\ref{chapter:training_dataset} as LUP-PS. To evaluate the generalization ability of our model, we conduct experiments on two benchmarks: CUHK-SYSU~\newcite{DBLP:conf/cvpr/XiaoLWLW17} and PRW~\newcite{DBLP:conf/cvpr/ZhengZSCYT17}. The CUHK-SYSU dataset comprises 18,184 images and 96,143 bounding boxes of 8,432 different identities, while the PRW dataset contains 11,816 images of 932 identities captured by six cameras.

\textbf{Evaluation Protocols}.
Under the domain generalization setting, all experiments in this study are conducted using LUP-PS as the training dataset, with the training sources partitioned into three distinct source domains based on the country label. All evaluations are performed on the test sets of CUHK-SYSU and PRW. Mean average precision (mAP) and cumulative matching characteristic (CMC) top-1 accuracy are employed as the evaluation metrics for the ReID sub-task, while average precision (AP) and recall rate are utilized as the metrics for the detection sub-task.

\begin{table*}[htbp]
\setlength{\abovecaptionskip}{2mm}
\centering
\caption{Comparative results when combining different designed schemes. \textbf{mDSBN} means multi-task protoype-based DSBN, \textbf{BR} refers to the Bounding boxes Refinement, \textbf{MLG} denotes the Missing label generation, \textbf{FD} means Feature Decorrelation, and \textbf{IE}, \textbf{ID} refers to Inter-frame Enhancement, Intra-frame Distinction respectively.}
\resizebox{2\columnwidth}{!}{
\begin{tabular}{cccccc|cccc|cccc}
\hline
         &   &  &     &     &     & \multicolumn{4}{c|}{Target: PRW}    & \multicolumn{4}{c}{Target: CUHK-SYSU}                        \\ \cline{7-14} 
 {\multirow{-2}{*}{mDSBN}}                  &  {\multirow{-2}{*}{BR}}            &  {\multirow{-2}{*}{MLG}}  &  {\multirow{-2}{*}{FD}} &  {\multirow{-2}{*}{IE}} &  {\multirow{-2}{*}{ID}}      & mAP     & top-1  & AP           & \multicolumn{1}{c|}{recall}                  & mAP     & top-1  &AP            & recall \\ 
\hline \hline
$\times$  &$\times$  &$\times$   &$\times$  & $\times$ & $\times$ & 12.7  & 64.3     & 71.4  & 72.3  & 75.6  & 78.6  & 63.4 &65.5  \\
$\checkmark$  &$\times$  &$\times$ & $\times$ & $\times$ & $\times$ &13.7  &67.0  &76.2  &78.0   &77.5  &80.3  &67.8  &70.8                \\
$\times$  &$\checkmark$  &$\times$ & $\times$ & $\times$ & $\times$ &13.2 &65.5 &80.1 &83.3 &76.1 &78.8 &72.9 &78.7\\
$\checkmark$ &$\checkmark$  &$\times$ & $\times$ & $\times$ & $\times$ &14.7  &70.0  &83.9  &88.1 &78.6  &79.9  &75.6 &81.6        \\ \hline
$\checkmark$ &$\checkmark$  &$\checkmark$ &$\times$ & $\times$ & $\times$ &16.1  &73.3  &87.8  &92.5  &80.4  &82.5  &\textbf{76.3}  &81.8        \\
$\checkmark$ &$\checkmark$  &$\times$ & $\checkmark$ & $\times$ & $\times$ &15.8 &73.1 &87.7 &92.6 &80.2 &82.3 &76.1 &82.1\\
$\checkmark$ &$\checkmark$  &$\checkmark$ & $\checkmark$ & $\times$ & $\times$  &16.6   & 73.9 & 87.3 & 93.3 &80.9 &83.1 & 75.8 &81.7                \\ \hline
$\checkmark$ &$\checkmark$  &$\checkmark$ & $\checkmark$ & $\checkmark$ & $\times$ &19.8  &76.6  &87.8  &94.1 &82.6  &83.9  &76.3  &82.1      \\
$\checkmark$ &$\checkmark$  &$\checkmark$ & $\checkmark$ & $\times$ & $\checkmark$ &20.3 &76.1 &87.6 &92.8 &81.4 &83.3 &76.2 &81.8\\
$\checkmark$ &$\checkmark$  &$\checkmark$ & $\checkmark$ & $\checkmark$ &$\checkmark$ &\textbf{21.9}  &\textbf{78.2}  &\textbf{88.0}  &\textbf{94.8}  &\textbf{84.1}  &\textbf{85.9}  &76.2 &\textbf{82.3} \\ \hline
\end{tabular}}
\label{tab:ablation}
\end{table*}

\textbf{Implementation Details}.
In accordance with our baseline framework~\newcite{DBLP:conf/aaai/LiM21}, we utilize a ResNet50 backbone that has been pre-trained on ImageNet to ensure a fair comparison. We set the momentum scalar $m$ to 0.9, and set a default margin $m_t$ of 0.3. The threshold $\psi$ for assigning a pseudo is set to 0.5. The input images $I$ are resized to 1500$\times$900 during the training process, while stochastic gradient descent (SGD) is applied for optimization over the course of 20 epochs using a batch size of 5. The learning rate is initially set to 0.003, with a decay factor of 0.1 being introduced after the twelfth epochs. The momentum and weight decay are set to 0.9 and $5 \times 10^{-4}$, respectively. Our model is trained with 8 NVIDIA A100 GPUs.

\subsection{Ablation Study}
In this section, we present a series of ablation experiments to validate the effectiveness of each proposed component. Table~\ref{tab:ablation} illustrates the results of combining different procedures, from which we observe that the direct transfer baseline method achieves 75.6\% mAP and 78.6\% top-1 accuracy on the CUHK-SYSU dataset. Subsequently, the addition of the multi-task prototype based DSBN and bounding box refinement techniques independently benefit both the ReID and detection sub-tasks. When these designs are combined, the proposed model outperforms the baseline by 3.0\% in mAP and 12.2\% in AP.
Furthermore, the proposed missing label generation and feature decorrelation methods lead to a 2.3\% increase in mAP and a 3.2\% improvement in top-1. The inter-frame enhancement and intra-frame distinction methods are designed to tackle the cross-camera problem and continually improve the performance by a large margin. By integrating all of these modules, our proposed method achieves 84.1\% mAP, 85.9\% top-1, and 76.2\% AP on the CUHK-SYSU dataset.
A similar improvement can be observed on PRW, with a 9.2\% increase in mAP, a 13.9\% improvement in top-1 accuracy. 

\begin{figure}[htbp]
    \setlength{\abovecaptionskip}{8pt}
    \centering
    \includegraphics[width=\linewidth]{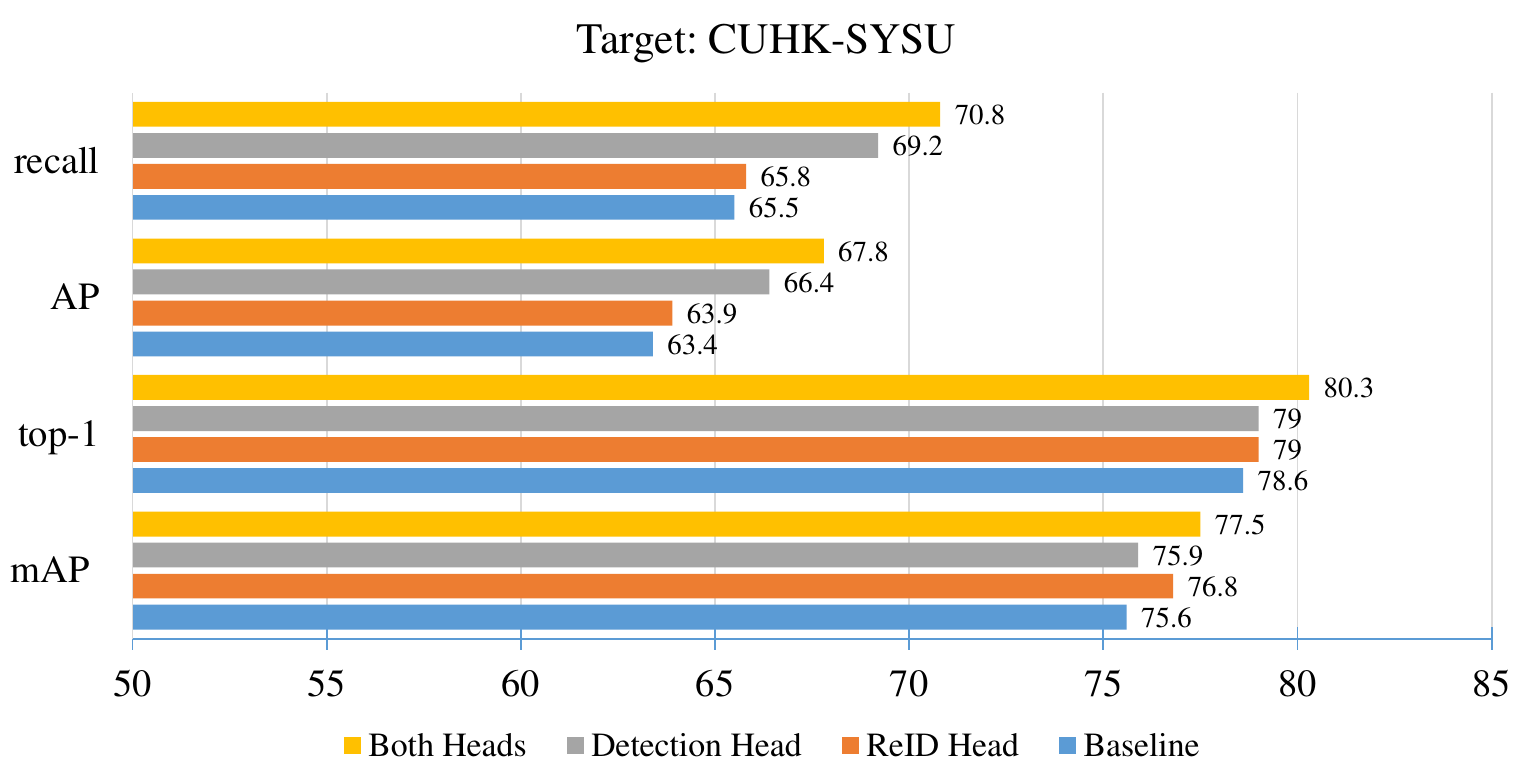}
    \caption{Results of where to apply prototype-based DSBN, `ReID Head' means using DSBN only in the ReID head, while `Detection Head' refers to only replacing all BN layers in the detection branch.}
    \label{fig:dsbn}
\end{figure}

\textbf{Effectiveness of mDSBN.}
The mDSBN maintains a prototype memory for each of the sub-task heads. We validate the effectiveness of applying DSBN for both sub-tasks and report the results in Figure~\ref{fig:dsbn}. The results indicate that the prototype-based DSBN improves the performance of the corresponding sub-task. Furthermore, by integrating these approaches, we achieve a further performance gain.

\begin{table}[htbp]
\setlength{\abovecaptionskip}{2mm}
\centering
\caption{Comparative results of different label generalization strategies, `Conventional' means directly assigning pseudo labels according to similarity to dientities.}
\begin{tabular}{p{2.5cm}|p{0.9cm}p{0.9cm}p{0.9cm}p{0.9cm}}
\hline
              & \multicolumn{4}{c}{Target: CUHK-SYSU}                        \\ \cline{2-5} 
{\multirow{-2}{*}{Label}}             & mAP     & top-1  & AP           & recall \\ 
\hline \hline  
Baseline   &78.6  &79.9  &75.6 &81.6 \\
Conventional  &79.6  &81.0   &75.9  &81.7\\
Hungarian   &\textbf{80.4}  &\textbf{82.5}  &\textbf{76.3}  &\textbf{81.8} \\\hline
\end{tabular}
\label{tab:pseudo}
\end{table}
\textbf{Effectiveness of Hungarian Algorithm.}
The process of generating missing labels in our work is accomplished through a bipartite matching formulation, which is subsequently solved using the Hungarian Algorithm. Our proposed approach is compared against the conventional label generation strategy, which directly assigns identity labels to unlabelled instances based on their highest similarity to identities within the same video. The comparison results, as presented in Table~\ref{tab:pseudo}, demonstrate the necessity of generating missing labels for open-world data and superiority of our bipartite matching formulation.

\begin{table}[htbp]
\setlength{\abovecaptionskip}{2mm}
\centering
\caption{Comparative results of decorrelation regularizer. `HSIC' refers to using HSIC~\newcite{DBLP:conf/nips/GrettonFTSSS07} to measure the feature independence, and `RF' refers to employing random features as kernel methods.}
\begin{tabular}{p{2.5cm}|p{0.9cm}p{0.9cm}|p{0.9cm}p{0.9cm}}
\hline
& \multicolumn{2}{c|}{Target: PRW} & \multicolumn{2}{c}{Target: CUHK}    \\ \cline{2-5}
{\multirow{-2}{*}{Strategy}}             & mAP     & top-1  & mAP     & top-1 \\ 
\hline \hline  
Baseline   &14.7  &70.0  &78.6  &79.9 \\
HSIC  &\textbf{16.0} &\textbf{73.1} &80.0 &81.6\\
RF &15.6 &72.8 &79.8 &81.0\\
Ours   &15.8 &\textbf{73.1}  &\textbf{80.2} &\textbf{82.3} \\\hline
\end{tabular}
\label{tab:decorrelation}
\end{table}
\textbf{Effectiveness of Feature Decorrelation.}
We compare different identity feature dimension decorrelation methods in Table~\ref{tab:decorrelation}. It can be observed that the decorrelation operation between ID-related dimensions and ID-irrelevant ones consistently improves discrimination performance. Our implementation can be viewed as an approximation to HSIC, but requires much less computation time. The experiment results in Table~\ref{tab:decorrelation} validate the reasonableness of using F-norm based regularizer as the replacement for HSIC. Moreover, we can observe that our implementation slightly outperforms random feature approximation.

\begin{table}[htbp]
\setlength{\abovecaptionskip}{2mm}
\centering
\caption{Comparing different strategies for selecting positive samples in the inter-frame loss. The `one-time memory' strategy utilizes the last seen instance of the same ID as the positive sample, while the `frame-id selected' strategy keeps track of all instances and selects those in frames more than 100 frames away from the anchor as positive samples.}
\begin{tabular}{p{2.5cm}|p{0.9cm}p{0.9cm}|p{0.9cm}p{0.9cm}}
\hline
& \multicolumn{2}{c|}{Target: PRW} & \multicolumn{2}{c}{Target: CUHK}    \\ \cline{2-5}
{\multirow{-2}{*}{Strategy}}             & mAP     & top-1  & mAP     & top-1 \\ 
\hline \hline  
Baseline   &16.6 &73.9  &80.9  &83.1 \\
One-time memory  &16.2 &73.9 &80.1 &82.3\\
Original prototype &17.1 &74.1 &81.3 &82.5\\
Frame-id selected &\textbf{20.2} &75.9  &\textbf{82.6} &\textbf{84.2}\\
Binary prototype   &19.8 &\textbf{76.6}  &\textbf{82.6} &83.9 \\\hline
\end{tabular}
\label{tab:inter-frame}
\end{table}
\textbf{Reasonableness of Inter-frame Enhancement.}
The primary challenge in the design of inter-frame enhancement (IE) lies in effectively constructing inter-frame positive samples for triplet training. In this study, we propose an effective and efficient strategy to address this issue. Our approach involves maintaining two prototypes for each identity, allowing us to separately represent instances from the first half and second half of the frame sequences. 
To assess the effectiveness of our implementation, we conducted a comprehensive comparison with several alternative positive sample selection strategies, as outlined in Table~\ref{tab:inter-frame}.

Specifically, we explored different approaches to obtain inter-frame samplers, which are described below.
(1) The `one-time memory' strategy involves recording the most recently encountered instance of each identity and utilizing it as the positive sample for subsequent instances.
(2) The `original prototype' approach directly employs the identity prototype stored in the memory bank as the inter-frame instance.
(3) The `frame-id selected' method keeps track of all instances associated with a particular identity. Given an anchor instance, this method selects instances from frames that are at least 100 frames away from the anchor, ensuring a sufficient temporal gap.
The results in Table~\ref{tab:inter-frame} indicate that the one-time memory strategy exhibited lower performance compared to the baseline, primarily due to its inherent instability. On the other hand, utilizing the prototypes directly from the memory bank as positive samples resulted in a slight improvement in identification performance. Since both the `one-time memory' and `original prototype' strategies strategies did not explicitly construct inter-frame triplets, the performance gains achieved were relatively modest. In contrast, when employing appropriate strategies to construct inter-frame triplets, we observed more significant improvements in performance. For example, our implementation outperforms the baseline by over 2\% mAP in PRW.
On the other hand, the `frame-id selected' strategy accurately provides same-identity samples from distant frames but incurs higher costs due to the need to store all instances. In contrast, our proposed `binary prototype' could be seen as a convenient and efficient alternative for maintaining inter-frame cases. Notably, our strategy demonstrates very similar performance to the frame-id based method while requiring only 6\% of the memory storage cost. These findings validate the effectiveness and reasonableness of the proposed inter-frame enhancement implementation. 
The IE method brings about 2\% mAP improvement on both datasets with very limited computation and storage cost. It is important to note that the IE module complements the ID module. As shown in Table~\ref{tab:ablation}, combining the IE and ID modules leads to further performance gains. This highlights the synergistic effect of integrating both modules and reinforces the value of our proposed inter-frame enhancement approach.


\begin{figure}[htbp]
\setlength{\abovecaptionskip}{8pt}
\centering
\subfloat[Target: PRW\label{subfig-1-1}]{%
   \includegraphics[width=0.7\linewidth]{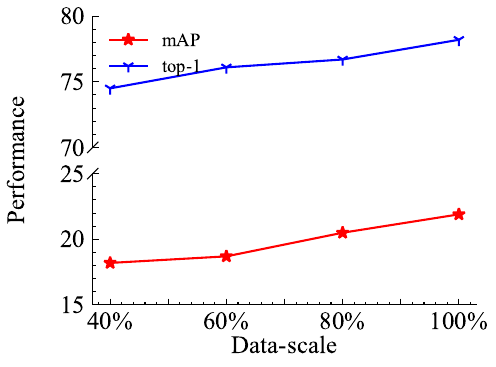}
}
\hspace{8mm}
\subfloat[Target: CUHK-SYSU\label{subfig-1-2}]{%
   \includegraphics[width=0.7\linewidth]{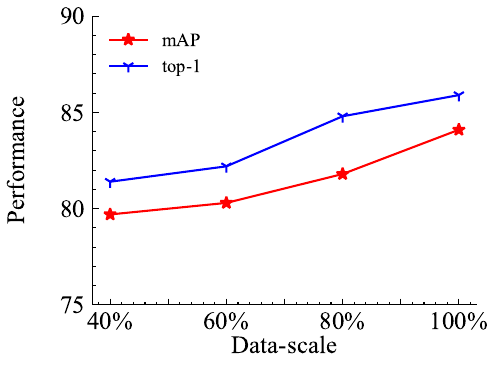}
}
\caption{The impact of open-world data scale on performance. All models are trained on the LUP-PS. We control the training data scale by randomly dropping part of source videos.}
 \label{fig:scale}
\end{figure}

\textbf{Influence of Data Scale.}
To account for training time considerations, we restricted the training data source to only three domains. In order to assess the impact of open-world video scale on training, we conducted experiments that randomly scaled down the data. The results, presented in Table~\ref{fig:scale}, indicate that the performance on both target datasets of the model improve as the training data-scale increases. Moreover, we observed that a larger scale of open-world training data does not lead to rapid convergence of generalization performance. This finding suggests that training generalizable person search models on a larger scale of open-world video frames holds promise for achieving better downstream performance in future research endeavors.

\textbf{Results with Different Base Models.}
\begin{table}[htbp]
\setlength{\abovecaptionskip}{2mm}
\centering
\caption{Comparative results of direct transfer performance of different supervised frameworks.}
\begin{tabular}{p{2.5cm}|p{0.9cm}p{0.9cm}|p{0.9cm}p{0.9cm}}
\hline
& \multicolumn{2}{c|}{Target: PRW} & \multicolumn{2}{c}{Target: CUHK}    \\ \cline{2-5}
{\multirow{-2}{*}{Strategy}}             & mAP     & top-1  & mAP     & top-1 \\ 
\hline \hline  
NAE   &11.2  &63.1  &72.4  &74.9 \\
PSTR  &9.5 &63.7 &60.6 &62.2\\
SeqNet & 12.7  & 64.3 & 75.6  & 78.6\\\hline
GPS(Ours)   &\textbf{21.9} &\textbf{78.2}  &\textbf{84.1} &\textbf{85.9} \\\hline
\end{tabular}
\label{tab:base}
\end{table}
To address the confounding effect of the underlying person search model architecture, we trained several base supervised models on LUP-PS and evaluated their direct transfer performance in Table~\ref{tab:base}. The results demonstrate that nearly all the supervised methods exhibit sub-optimal performance when directly applied to the generalization scenario. An interesting observation is that some recently proposed state-of-the-art (SOTA) model architectures, such as PSTR~\newcite{DBLP:conf/cvpr/CaoPAC0SK22}, underperform prior approaches in terms of generalization ability. This phenomenon indicates the higher overfitting tendency of these advanced architectures. On the contrary, our proposed generalizable person search (GPS) framework outperforms the extended supervised methods by a significant margin. It is worth noting that the GPS framework is model-agnostic and can be easily integrated with other person search frameworks.

\begin{table}[htbp]
\setlength{\abovecaptionskip}{2mm}
\centering
\caption{Results of mixing open-world UGC video content and surveillance data as the training source. All experiments in this table follow the cross-domain evaluation,~\emph{i.e.}, the training set and the test set belong to different domains. }
\begin{tabular}{p{2.5cm}|p{0.9cm}p{0.9cm}|p{0.9cm}p{0.9cm}}
\hline
& \multicolumn{2}{c|}{Target: PRW} & \multicolumn{2}{c}{Target: CUHK}    \\ \cline{2-5}
{\multirow{-2}{*}{Training Sources}}            & mAP     & top-1  & mAP     & top-1 \\ 
\hline \hline 
CUHK-SYSU  &30.3 &77.7 &- &-\\
PRW &- &- &52.5 &54.8 \\
UGC Videos   &21.9 &78.2  &84.1 &85.9 \\
UGC + CUHK  &\textbf{32.1} &\textbf{82.2} &- &- \\
UGC + PRW   &- &- &\textbf{86.9} &\textbf{88.4} \\\hline
\end{tabular}
\label{tab:mix}
\end{table}
\textbf{Mixed-Sources Training Results.}
While the experiments conducted so far have demonstrated the effectiveness of utilizing open-world videos for training person search models, it is important to acknowledge the existence of some widely used public surveillance datasets. It is both reasonable and practical to combine open-world videos with available surveillance data as a hybrid training source. In this setting, we combine our open-world UGC data with one of the public surveillance datasets, such as CUHK-SYSU or PRW, for training, while leaving the other dataset as the target domain.
The results of training on mixed-sources datasets is reported in table~\ref{tab:mix}. It is evident that incorporating open-world videos into the training process improves stability and significantly enhances the identification ability across various surveillance scenes. Additionally, incorporating available surveillance data helps mitigate the issue of lacking cross-camera training samples in open-world videos. By combining these two sources, our framework demonstrates a significant advantage in generalization and outperforms almost all adaptive and weakly-supervised methods, without requiring any samples from the target domain.

\subsection{Comparison with State-of-the-Art}
\begin{table}[htbp]
\caption{Comparison with state-of-the-art methods. Fully supervised and weakly supervised methods employ the training data from the same domain as the test set. Domain adaptive approaches use training data from both datasets. In contrast, our domain generalization method does not employ any samples from the target domain.}
\label{tab:sota}
\centering
\resizebox{1\columnwidth}{!}{
\begin{tabular}{c|c|cc|cc}
\hline
\multicolumn{2}{c}{\multirow{2}*{Methods}}&
\multicolumn{2}{|c|}{CUHK-SYSU}&\multicolumn{2}{c}{PRW}\cr
\cline{3-6} 
\multicolumn{2}{c|}{}&mAP&top-1&mAP&top-1\cr
\hline
\multirow{9}* {\rotatebox{90}{Two-step}}
&\multicolumn{5}{l}{\textit{Fully Supervised: }}\cr
\cline{2-6} 
&IDE \newcite{DBLP:conf/cvpr/ZhengZSCYT17}&-&-&20.5&48.3\cr
&MGTS \newcite{DBLP:conf/eccv/ChenZOYT18}&83.0&83.7&32.6&72.1\cr
&IGPN \newcite{DBLP:conf/cvpr/DongZST20}&90.3&91.4&47.2&87.0\cr
&TCTS \newcite{DBLP:conf/cvpr/WangMCSC20}&93.9&95.1&46.8&87.5\cr
\cline{2-6} 
&\multicolumn{5}{l}{\textit{Weakly Supervised: }}\cr
\cline{2-6} 
&CGUA \newcite{DBLP:journals/corr/abs-2203-14307}&91.0&92.2&42.7&86.9\cr
\hline
\hline
\multirow{22}*{\rotatebox{90}{One-step}}
&\multicolumn{5}{l}{\textit{Fully Supervised: }}\cr
\cline{2-6} 
&OIM \newcite{DBLP:conf/cvpr/XiaoLWLW17}&75.5&78.7&21.3&49.4\\
&NPSM \newcite{DBLP:conf/iccv/LiuFJKZQJY17}&77.9&81.2&24.2&53.1\cr
&RCAA \newcite{DBLP:conf/eccv/ChangHSLYH18}&79.3&81.3&-&-\cr
&CTXG \newcite{DBLP:conf/cvpr/YanZNZXY19}&84.1&86.5&33.4&73.6\cr
&NAE \newcite{DBLP:journals/ijcv/ChenZYS21}&91.5&92.4&43.3&80.9\cr
&SeqNet \newcite{DBLP:conf/aaai/LiM21}&93.8&94.6&46.7&83.4\cr
&OIMNet++\newcite{DBLP:conf/eccv/LeeOBLH22}&93.1&94.1&47.7&84.8\cr
&PSTR \newcite{DBLP:conf/cvpr/CaoPAC0SK22}&93.5&95.0&49.5&87.8\cr
&COAT \newcite{DBLP:conf/cvpr/YuDLDFHC22}&94.2&94.7&53.3&87.4\cr
\cline{2-6} 
&\multicolumn{5}{l}{\textit{Weakly Supervised: }}\cr
\cline{2-6} 
&CGPS \newcite{DBLP:conf/aaai/YanLLQNLY22}&80.0&82.3&16.2&68.0\cr
&R-SiamNet \newcite{DBLP:conf/iccv/HanSYYGS0W21}&86.0&87.1&21.2&73.4\cr
\cline{2-6} 
&\multicolumn{5}{l}{\textit{UDA: }}\cr
\cline{2-6} 
&DAPS \newcite{DBLP:conf/eccv/LiYWYJD22} &77.6 &79.6 &34.7 & 80.6\cr
\cline{2-6} 
&\multicolumn{5}{l}{\textit{DG: }}\cr
\cline{2-6}
&Baseline &75.6 &78.6 &12.7 & 64.3\cr
&GPS (Ours, w/o surveillance data) &84.1 &85.9 &21.9 & 78.2\cr
&GPS (Ours, w/ surveillance data) &86.9 &88.4 &32.1 & 82.2\cr
\hline
\end{tabular}}
\end{table}
Based on the fact that the proposed framework is the first to explore generalizable person search with large-scale UGC data, we compared its performance with State-of-the-Art fully supervised, weakly supervised, and domain adaptation frameworks in Table~\ref{tab:sota}. The huge domain gap between UGC and surveillance data is the reason for the unsatisfactory baseline performance, and our proposed GPS framework leads the baseline by a large margin due to better generalization ability. The results indicate that our method not only significantly outperforms the baseline but also achieves a surprising improvement over many existing approaches. For instance, on the CUHK-SYSU dataset, our method outperforms several fully supervised~\newcite{DBLP:conf/cvpr/XiaoLWLW17,DBLP:journals/pr/XiaoXTHWF19,DBLP:conf/iccv/LiuFJKZQJY17,DBLP:conf/eccv/ChangHSLYH18}, weakly supervised~\newcite{DBLP:conf/aaai/YanLLQNLY22}, and domain adaptation~\newcite{DBLP:conf/eccv/LiYWYJD22} approaches. These results encourage further exploration to improve the generalization ability for arbitrary scenes.

\begin{figure}[t]
    \setlength{\abovecaptionskip}{8pt}
    \centering
    \includegraphics[width=\linewidth]{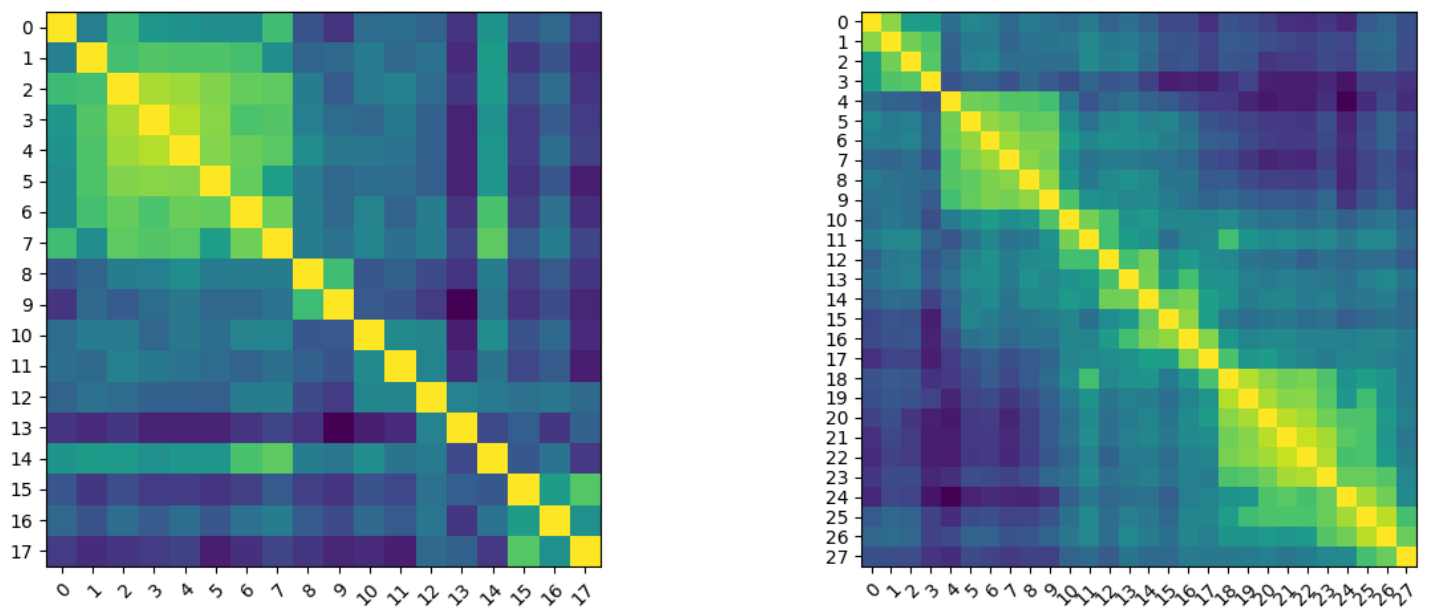}
    \caption{Similarity matrix of samples from different frames belonging to the same identity. The instances belonging to the same identity are arranged according to the order of the frame indexes they appear. Please note that the model for this visualization is trained without the inter-frame enhancement strategy.}
    \label{fig:similarity}
\end{figure}
\begin{figure}[t]
    \setlength{\abovecaptionskip}{8pt}
    \centering
    \includegraphics[width=\linewidth]{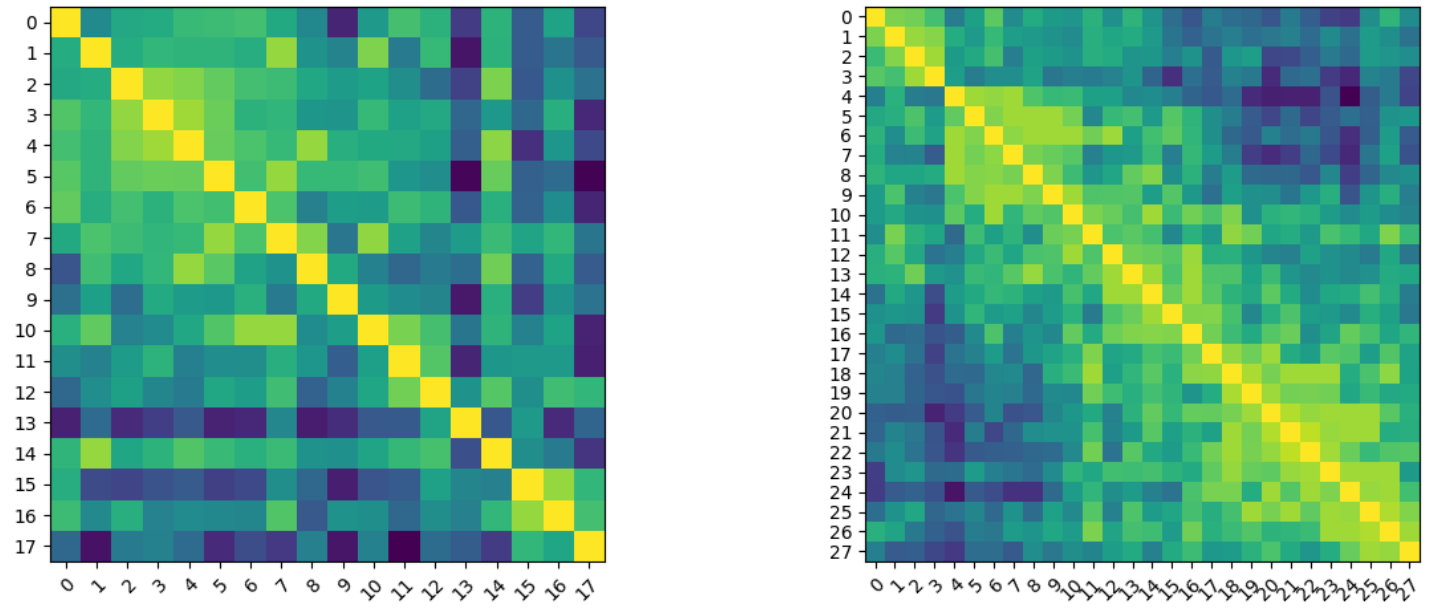}
    \caption{After introducing the inter-frame enhancement to training, the similarity matrix of samples from the same identities as Figure~\ref{fig:similarity}.}
    \label{fig:similarity_after}
\end{figure}
\begin{figure*}[htbp]
    \setlength{\abovecaptionskip}{8pt}
    \centering
    \includegraphics[width=0.9\linewidth]{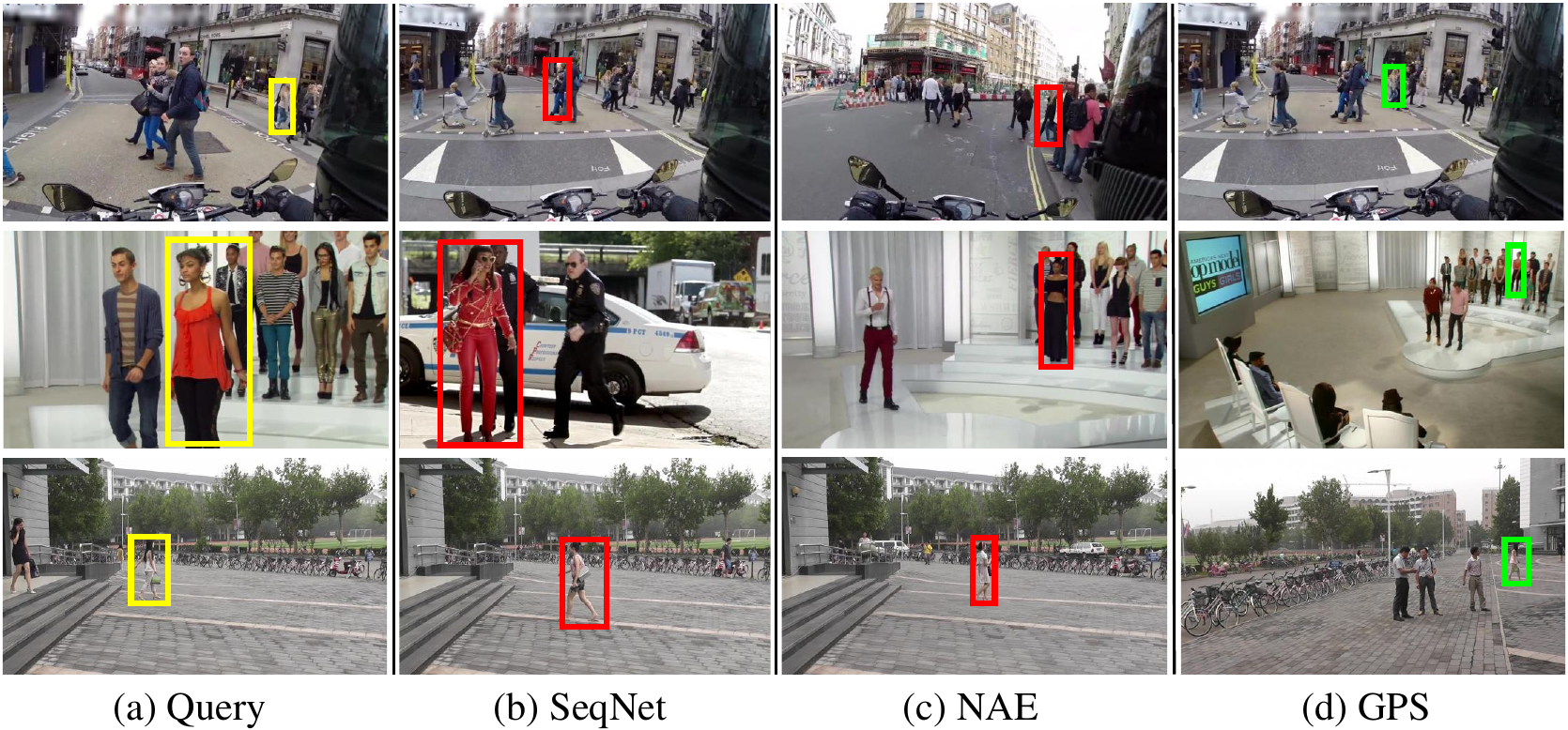}
    \caption{Visualization of difficult cases that successfully retrieved by our proposed GPS, but not the baseline methods. The yellow boxes refer to the queries, while the green and red boxes denote the correct and incorrect top-1 retrieving results, respectively. All the models are trained with the LUP-PS dataset.}
    \label{fig:visual}
\end{figure*}
\begin{figure*}[htbp]
    \setlength{\abovecaptionskip}{8pt}
    \centering
    \includegraphics[width=0.9\linewidth]{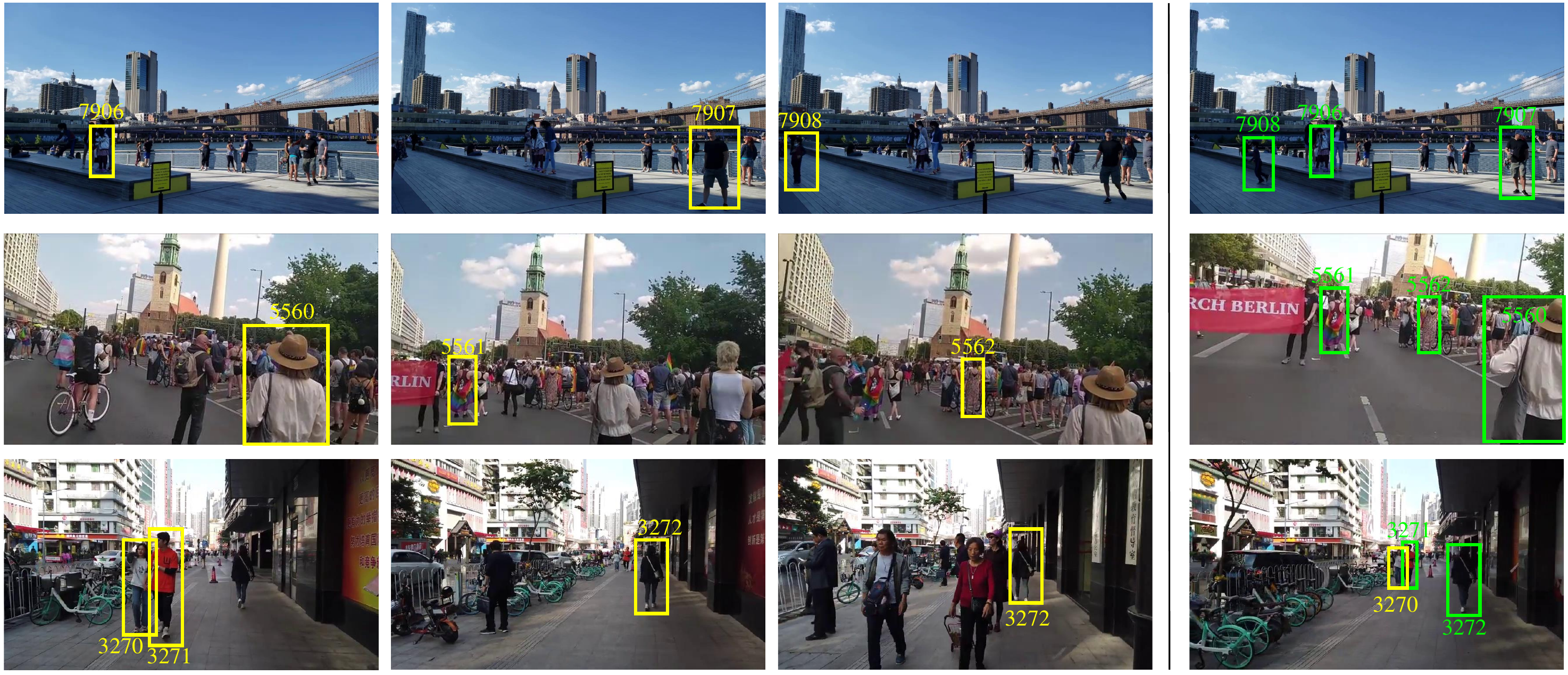}
    \caption{Visualization of label generation results, the left three columns are example frames with original annotation, and the right column is the results of label generation. Yellow bounding boxes refer to original annotation, and green boxes denote the generated labels.}
    \label{fig:assign}
\end{figure*}

\subsection{Discussions}
\label{discussion}
\textbf{Discussion on Comparison with DA.}
Domain adaptation (DA) and domain generalization (DG) are two common approaches to applying a model to novel scenes. However, DA models need to re-train the model to adapt to novel scenes, and they require different models for each scene. This is time-consuming and memory-inefficient. For person search, DA models have been explored, \emph{i.e.}, DAPS~\newcite{DBLP:conf/eccv/LiYWYJD22}, but it still requires over a day to re-train the model on a PRW-scale dataset. In this work, we introduce the DG setting to person search, enabling a single model and one-time training for all the application scenarios. Our framework not only exhibits significant advantages over DA methods in terms of rapid deployment and memory efficiency, but also achieves better performance in some critical scenarios, such as CUHK-SYSU.


\textbf{Discussion on Missing Label Generation.}
To address the label omission problem, we propose assigning pseudo identity labels to qualified None-ID instances and frame this task as a bipartite matching problem. Although there are alternative solutions available, such as generating pseudo labels for all instances in a video source through clustering, we have chosen the assignment strategy for two primary reasons. Firstly, in our setting, clustering needs to be performed for each video at every epoch, which incurs a significant time cost. Specifically, the clustering method takes approximately 20 hours per epoch, whereas our implementation of the assignment strategy takes only 20 minutes per epoch. Secondly, existing identity labels derived from tracklets are relatively reliable, whereas the clustering strategy ignores this information and may result in unstable outcomes during the initial stages of training.

\textbf{Discussion on Inter-frame Enhancement.}
In this work, we leverage binary prototypes as positive samples for inter-frame enhancement, and we provide a theoretical analysis of the rationale behind our implementation. To do so, we first define `inter-frame' instances as those whose distance to an anchor feature $\hat{x}$ is below a threshold $\psi=0.6$, the same as the similarity threshold in label generation. 
\begin{equation}
\operatorname{cos}(\hat{x}, x) < \psi,
\end{equation}
Under this definition, we can infer that the sampling probability follows the probability distribution of inter-frame instances given an anchor feature $x$. Empirically, the similarity between two frames decreases and gradually converges to stability as the frames become further apart. A Gaussian-like distribution density function is a common formulation to capture this trend.
\begin{equation}
g(x)=\frac{1}{\sigma \sqrt{2 \pi}} e^{-\frac{(x-\hat{x})^2}{2 \sigma^2}}+b.
\end{equation}
To validate our hypothesis, we analyze the similarity distributions of over 30 identities from randomly picked videos beyond the training set. Some visualization results are presented in Figure~\ref{fig:similarity}. The results show that the similarity heat-maps align with our hypothesis, with the interval between $\hat{x}$ and $x_{ie}$ being 10.
Recalling that we update of prototypes with a momentum-based method in formulation~\ref{formulation:update},
where $m$ is the momentum factor and $c$ represents the current identity prototype. By expanding $c$ into a series, we obtain:
\begin{equation}
c = \sum \limits_{{i=1}}^{{ n }} m^{i-1}(1-m) \hat{x}^{-i} ,
\end{equation}
where $\hat{x}^{-i}$ denotes the $i$-th previous instance with the same identity as $\hat{x}$. Since all $\hat{x}^{-i}$ follow the inter-frame similarity distribution of $\hat{x}$, and $c$ is a linear combination of them. Considering that each identity is distributed over an average of 30 non-contiguous frames after label generation, the probability of $c$ being an inter-frame instance is nearly 0.85. After applying our IE design, a notable improvement in the similarity between instances from far-away frames can be observed in Figure~\ref{fig:similarity_after}. This further validates the rationale behind our implementation.

\subsection{Qualitative Results}
We present the visualization of our retrieval results in Figure~\ref{fig:visual}. Our proposed GPS framework demonstrates its efficacy in accurately identifying the target identity even in complex scenarios, such as crowded and cross-camera scenarios. The qualitative analysis of the results provides compelling evidence for the effectiveness of the learning strategies incorporated into our framework.

We also demonstrate some visualization results in Figure~\ref{fig:assign} to validate the correctness of the proposed label generation. We can infer from Figure~\ref{fig:assign} that almost all existing identities are properly assigned owing to the Hungarian algorithm, and these results fully demonstrate the feasibility and effectiveness of pseudo label assignment.

\section{Conclusion}
In this work, we make the first step towards training person search models using large-scale open-world UGC videos without human-crafted curation, and present a novel and practical generalizable person search setting. Based on our proposed  setting, we develop a strong baseline by exploring feature-level invariant learning and data-level generalization methods. The experiment results validate the superiority of our framework and the feasibility of using open-world videos for training, and we sincerely hope this work will encourage more exploration towards solving the generalizable person search problem.


{\footnotesize
\bibliographystyle{spbasic}
\bibliography{egbib}

\begin{thebibliography}{62}
\providecommand{\natexlab}[1]{#1}
\providecommand{\url}[1]{{#1}}
\providecommand{\urlprefix}{URL }
\expandafter\ifx\csname urlstyle\endcsname\relax
  \providecommand{\doi}[1]{DOI~\discretionary{}{}{}#1}\else
  \providecommand{\doi}{DOI~\discretionary{}{}{}\begingroup
  \urlstyle{rm}\Url}\fi
\providecommand{\eprint}[2][]{\url{#2}}

\bibitem[{Baker(1973)}]{baker1973joint}
Baker CR (1973) Joint measures and cross-covariance operators. Transactions of
  the American Mathematical Society pp 273--289

\bibitem[{Balaji et~al.(2018)Balaji, Sankaranarayanan, and
  Chellappa}]{DBLP:conf/nips/BalajiSC18}
Balaji Y, Sankaranarayanan S, Chellappa R (2018) Metareg: Towards domain
  generalization using meta-regularization. In: NeurIPS, pp 1006--1016

\bibitem[{Cao et~al.(2022)Cao, Pang, Anwer, Cholakkal, Xie, Shah, and
  Khan}]{DBLP:conf/cvpr/CaoPAC0SK22}
Cao J, Pang Y, Anwer RM, Cholakkal H, Xie J, Shah M, Khan FS (2022) {PSTR:}
  end-to-end one-step person search with transformers. In: {CVPR}, pp
  9448--9457

\bibitem[{Chang et~al.(2019)Chang, You, Seo, Kwak, and
  Han}]{DBLP:conf/cvpr/ChangYSKH19}
Chang W, You T, Seo S, Kwak S, Han B (2019) Domain-specific batch normalization
  for unsupervised domain adaptation. In: {CVPR}, pp 7354--7362

\bibitem[{Chang et~al.(2018)Chang, Huang, Shen, Liang, Yang, and
  Hauptmann}]{DBLP:conf/eccv/ChangHSLYH18}
Chang X, Huang P, Shen Y, Liang X, Yang Y, Hauptmann AG (2018) {RCAA:}
  relational context-aware agents for person search. In: Ferrari V, Hebert M,
  Sminchisescu C, Weiss Y (eds) {ECCV}, pp 86--102

\bibitem[{Chen et~al.(2018)Chen, Zhang, Ouyang, Yang, and
  Tai}]{DBLP:conf/eccv/ChenZOYT18}
Chen D, Zhang S, Ouyang W, Yang J, Tai Y (2018) Person search via a mask-guided
  two-stream {CNN} model. In: {ECCV}, pp 764--781

\bibitem[{Chen et~al.(2020)Chen, Zhang, Ouyang, Yang, and
  Schiele}]{DBLP:conf/aaai/ChenZO0S20}
Chen D, Zhang S, Ouyang W, Yang J, Schiele B (2020) Hierarchical online
  instance matching for person search. In: {AAAI}, pp 10518--10525

\bibitem[{Chen et~al.(2021)Chen, Zhang, Yang, and
  Schiele}]{DBLP:journals/ijcv/ChenZYS21}
Chen D, Zhang S, Yang J, Schiele B (2021) Norm-aware embedding for efficient
  person search and tracking. Int J Comput Vis 129(11):3154--3168

\bibitem[{Chen and Xu(2022)}]{DBLP:journals/corr/abs-2211-04323}
Chen L, Xu J (2022) Sequential transformer for end-to-end person search. CoRR
  abs/2211.04323

\bibitem[{Choi et~al.(2021)Choi, Kim, Jeong, Park, and
  Kim}]{DBLP:conf/cvpr/ChoiKJPK21}
Choi S, Kim T, Jeong M, Park H, Kim C (2021) Meta batch-instance normalization
  for generalizable person re-identification. In: {CVPR}, pp 3425--3435

\bibitem[{Dai et~al.(2021)Dai, Li, Liu, Tong, and
  Duan}]{DBLP:conf/cvpr/DaiLLTD21}
Dai Y, Li X, Liu J, Tong Z, Duan L (2021) Generalizable person
  re-identification with relevance-aware mixture of experts. In: {CVPR}, pp
  16145--16154

\bibitem[{Dong et~al.(2020)Dong, Zhang, Song, and
  Tan}]{DBLP:conf/cvpr/DongZST20}
Dong W, Zhang Z, Song C, Tan T (2020) Instance guided proposal network for
  person search. In: {CVPR}, pp 2582--2591

\bibitem[{Fu et~al.(2021)Fu, Chen, Bao, Yang, Yuan, Zhang, Li, and
  Chen}]{DBLP:conf/cvpr/Fu0BYY0L021}
Fu D, Chen D, Bao J, Yang H, Yuan L, Zhang L, Li H, Chen D (2021) Unsupervised
  pre-training for person re-identification. In: {CVPR}, pp 14750--14759

\bibitem[{Fu et~al.(2022)Fu, Chen, Yang, Bao, Yuan, Zhang, Li, Wen, and
  Chen}]{DBLP:conf/cvpr/Fu0YBY0LW022}
Fu D, Chen D, Yang H, Bao J, Yuan L, Zhang L, Li H, Wen F, Chen D (2022)
  Large-scale pre-training for person re-identification with noisy labels. In:
  {CVPR}, pp 1--11

\bibitem[{Gretton et~al.(2005)Gretton, Herbrich, Smola, Bousquet, and
  Sch{\"{o}}lkopf}]{DBLP:journals/jmlr/GrettonHSBS05}
Gretton A, Herbrich R, Smola AJ, Bousquet O, Sch{\"{o}}lkopf B (2005) Kernel
  methods for measuring independence. J Mach Learn Res pp 2075--2129

\bibitem[{Gretton et~al.(2007)Gretton, Fukumizu, Teo, Song, Sch{\"{o}}lkopf,
  and Smola}]{DBLP:conf/nips/GrettonFTSSS07}
Gretton A, Fukumizu K, Teo CH, Song L, Sch{\"{o}}lkopf B, Smola AJ (2007) A
  kernel statistical test of independence. In: {NIPS}, pp 585--592

\bibitem[{Han et~al.(2021{\natexlab{a}})Han, Ko, and
  Sim}]{DBLP:conf/bmvc/HanKS21}
Han B, Ko K, Sim J (2021{\natexlab{a}}) Context-aware unsupervised clustering
  for person search. In: {BMVC}, p 386

\bibitem[{Han et~al.(2021{\natexlab{b}})Han, Su, Yu, Yuan, Gao, Sang, Yang, and
  Wang}]{DBLP:conf/iccv/HanSYYGS0W21}
Han C, Su K, Yu D, Yuan Z, Gao C, Sang N, Yang Y, Wang C (2021{\natexlab{b}})
  Weakly supervised person search with region siamese networks. In: {ICCV}, pp
  11986--11995

\bibitem[{Jaffe and Zakhor(2023)}]{DBLP:conf/wacv/JaffeZ23}
Jaffe L, Zakhor A (2023) Gallery filter network for person search. In: {WACV},
  {IEEE}, pp 1684--1693

\bibitem[{Jia et~al.(2022)Jia, Luo, Yan, Chang, and
  Zheng}]{DBLP:journals/corr/abs-2203-14307}
Jia C, Luo M, Yan C, Chang X, Zheng Q (2022) {CGUA:} context-guided and
  unpaired-assisted weakly supervised person search. CoRR abs/2203.14307

\bibitem[{Jin et~al.(2020)Jin, Lan, Zeng, Chen, and
  Zhang}]{DBLP:conf/cvpr/JinLZ0Z20}
Jin X, Lan C, Zeng W, Chen Z, Zhang L (2020) Style normalization and
  restitution for generalizable person re-identification. In: {CVPR}, pp
  3140--3149

\bibitem[{Kuhn(1955)}]{kuhn1955hungarian}
Kuhn HW (1955) The hungarian method for the assignment problem. Naval research
  logistics quarterly pp 83--97

\bibitem[{Lee et~al.(2022)Lee, Oh, Baek, Lee, and
  Ham}]{DBLP:conf/eccv/LeeOBLH22}
Lee S, Oh Y, Baek D, Lee J, Ham B (2022) Oimnet++: Prototypical normalization
  and localization-aware learning for person search. In: Avidan S, Brostow GJ,
  Ciss{\'{e}} M, Farinella GM, Hassner T (eds) {ECCV}, pp 621--637

\bibitem[{Li et~al.(2018)Li, Yang, Song, and
  Hospedales}]{DBLP:conf/aaai/LiYSH18}
Li D, Yang Y, Song Y, Hospedales TM (2018) Learning to generalize:
  Meta-learning for domain generalization. In: {AAAI}, pp 3490--3497

\bibitem[{Li et~al.(2022)Li, Yan, Wang, Yu, Jia, and
  Ding}]{DBLP:conf/eccv/LiYWYJD22}
Li J, Yan Y, Wang G, Yu F, Jia Q, Ding S (2022) Domain adaptive person search.
  In: {ECCV}, pp 302--318

\bibitem[{Li and Miao(2021)}]{DBLP:conf/aaai/LiM21}
Li Z, Miao D (2021) Sequential end-to-end network for efficient person search.
  In: {AAAI}, pp 2011--2019

\bibitem[{Liao and Shao(2020)}]{DBLP:conf/eccv/LiaoS20}
Liao S, Shao L (2020) Interpretable and generalizable person re-identification
  with query-adaptive convolution and temporal lifting. In: {ECCV}, pp 456--474

\bibitem[{Liao and Shao(2022)}]{DBLP:conf/cvpr/Liao022}
Liao S, Shao L (2022) Graph sampling based deep metric learning for
  generalizable person re-identification. In: {CVPR}, {IEEE}, pp 7349--7358

\bibitem[{Lin et~al.(2021)Lin, Li, and Kot}]{DBLP:journals/tip/LinLK21}
Lin S, Li C, Kot AC (2021) Multi-domain adversarial feature generalization for
  person re-identification. {IEEE} Trans Image Process pp 1596--1607

\bibitem[{Liu et~al.(2017)Liu, Feng, Jie, Karlekar, Zhao, Qi, Jiang, and
  Yan}]{DBLP:conf/iccv/LiuFJKZQJY17}
Liu H, Feng J, Jie Z, Karlekar J, Zhao B, Qi M, Jiang J, Yan S (2017) Neural
  person search machines. In: {ICCV}, pp 493--501

\bibitem[{Liu et~al.(2022)Liu, Huang, Li, Zheng, and Zha}]{dbn}
Liu J, Huang Z, Li L, Zheng K, Zha Z (2022) Debiased batch normalization via
  gaussian process for generalizable person re-identification. In: {AAAI}

\bibitem[{Muandet et~al.(2013)Muandet, Balduzzi, and
  Sch{\"{o}}lkopf}]{DBLP:conf/icml/MuandetBS13}
Muandet K, Balduzzi D, Sch{\"{o}}lkopf B (2013) Domain generalization via
  invariant feature representation. In: {ICML}, pp 10--18

\bibitem[{Munjal et~al.(2019)Munjal, Amin, Tombari, and
  Galasso}]{DBLP:conf/cvpr/MunjalATG19}
Munjal B, Amin S, Tombari F, Galasso F (2019) Query-guided end-to-end person
  search. In: {CVPR}, pp 811--820

\bibitem[{Noura et~al.(2021)Noura, Salman, Couturier, and
  Sider}]{DBLP:journals/corr/abs-2104-13634}
Noura HN, Salman O, Couturier R, Sider A (2021) A deep learning object
  detection method for an efficient clusters initialization. CoRR
  abs/2104.13634

\bibitem[{Pu et~al.(2023)Pu, Zhong, Sebe, and Lew}]{DBLP:journals/pami/PuZSL23}
Pu N, Zhong Z, Sebe N, Lew MS (2023) A memorizing and generalizing framework
  for lifelong person re-identification. IEEE Trans Pattern Anal Mach Intell
  45(11):13567--13585

\bibitem[{Rahimi and Recht(2007)}]{DBLP:conf/nips/RahimiR07}
Rahimi A, Recht B (2007) Random features for large-scale kernel machines. In:
  {NIPS}, pp 1177--1184

\bibitem[{Song et~al.(2019)Song, Yang, Song, Xiang, and
  Hospedales}]{DBLP:conf/cvpr/SongYSXH19}
Song J, Yang Y, Song Y, Xiang T, Hospedales TM (2019) Generalizable person
  re-identification by domain-invariant mapping network. In: {CVPR}, pp
  719--728

\bibitem[{Song et~al.(2012)Song, Smola, Gretton, Bedo, and
  Borgwardt}]{DBLP:journals/jmlr/SongSGBB12}
Song L, Smola AJ, Gretton A, Bedo J, Borgwardt KM (2012) Feature selection via
  dependence maximization. J Mach Learn Res pp 1393--1434

\bibitem[{Strobl et~al.(2019)Strobl, Zhang, and
  Visweswaran}]{strobl2019approximate}
Strobl EV, Zhang K, Visweswaran S (2019) Approximate kernel-based conditional
  independence tests for fast non-parametric causal discovery. Journal of
  Causal Inference (1)

\bibitem[{Vidit et~al.(2023)Vidit, Engilberge, and
  Salzmann}]{DBLP:conf/cvpr/ViditES23}
Vidit V, Engilberge M, Salzmann M (2023) {CLIP} the gap: {A} single domain
  generalization approach for object detection. In: {CVPR}, {IEEE}, pp
  3219--3229

\bibitem[{Volpi and Murino(2019)}]{DBLP:conf/iccv/VolpiM19}
Volpi R, Murino V (2019) Addressing model vulnerability to distributional
  shifts over image transformation sets. In: {ICCV}, pp 7979--7988

\bibitem[{Volpi et~al.(2018)Volpi, Namkoong, Sener, Duchi, Murino, and
  Savarese}]{DBLP:conf/nips/VolpiNSDMS18}
Volpi R, Namkoong H, Sener O, Duchi JC, Murino V, Savarese S (2018)
  Generalizing to unseen domains via adversarial data augmentation. In:
  NeurIPS, pp 5339--5349

\bibitem[{Wang et~al.(2020)Wang, Ma, Chang, Shan, and
  Chen}]{DBLP:conf/cvpr/WangMCSC20}
Wang C, Ma B, Chang H, Shan S, Chen X (2020) {TCTS:} {A} task-consistent
  two-stage framework for person search. In: {CVPR}, pp 11949--11958

\bibitem[{Xiao et~al.(2019)Xiao, Xie, Tillo, Huang, Wei, and
  Feng}]{DBLP:journals/pr/XiaoXTHWF19}
Xiao J, Xie Y, Tillo T, Huang K, Wei Y, Feng J (2019) {IAN:} the individual
  aggregation network for person search. Pattern Recognit pp 332--340

\bibitem[{Xiao et~al.(2017)Xiao, Li, Wang, Lin, and
  Wang}]{DBLP:conf/cvpr/XiaoLWLW17}
Xiao T, Li S, Wang B, Lin L, Wang X (2017) Joint detection and identification
  feature learning for person search. In: {CVPR}, pp 3376--3385

\bibitem[{Xu et~al.(2022)Xu, Liang, He, and Sun}]{DBLP:conf/eccv/XuLHS22}
Xu B, Liang J, He L, Sun Z (2022) Mimic embedding via adaptive aggregation:
  Learning generalizable person re-identification. In: {ECCV}, pp 372--388

\bibitem[{Yan et~al.(2019)Yan, Zhang, Ni, Zhang, Xu, and
  Yang}]{DBLP:conf/cvpr/YanZNZXY19}
Yan Y, Zhang Q, Ni B, Zhang W, Xu M, Yang X (2019) Learning context graph for
  person search. In: {CVPR}, pp 2158--2167

\bibitem[{Yan et~al.(2021)Yan, Li, Liao, Qin, Ni, and
  Yang}]{DBLP:journals/corr/abs-2111-14290}
Yan Y, Li J, Liao S, Qin J, Ni B, Yang X (2021) {TAL:} two-stream adaptive
  learning for generalizable person re-identification. CoRR abs/2111.14290

\bibitem[{Yan et~al.(2022)Yan, Li, Liao, Qin, Ni, Lu, and
  Yang}]{DBLP:conf/aaai/YanLLQNLY22}
Yan Y, Li J, Liao S, Qin J, Ni B, Lu K, Yang X (2022) Exploring visual context
  for weakly supervised person search. In: {AAAI}, pp 3027--3035

\bibitem[{Yan et~al.(2023)Yan, Li, Qin, Zheng, Liao, and
  Yang}]{DBLP:journals/ijcv/YanLQZLY23}
Yan Y, Li J, Qin J, Zheng P, Liao S, Yang X (2023) Efficient person search: An
  anchor-free approach. Int J Comput Vis 131(7):1642--1661

\bibitem[{Yu et~al.(2022)Yu, Du, LaLonde, Davila, Funk, Hoogs, and
  Clipp}]{DBLP:conf/cvpr/YuDLDFHC22}
Yu R, Du D, LaLonde R, Davila D, Funk C, Hoogs A, Clipp B (2022) Cascade
  transformers for end-to-end person search. In: {CVPR}, pp 7257--7266

\bibitem[{Yuan et~al.(2023)Yuan, Ma, Chen, Kuang, Wu, and
  Lin}]{DBLP:journals/ijcv/YuanMCKWL23}
Yuan J, Ma X, Chen D, Kuang K, Wu F, Lin L (2023) Domain-specific bias
  filtering for single labeled domain generalization. Int J Comput Vis
  131(2):552--571

\bibitem[{Yue et~al.(2019)Yue, Zhang, Zhao, Sangiovanni{-}Vincentelli, Keutzer,
  and Gong}]{DBLP:conf/iccv/YueZZSKG19}
Yue X, Zhang Y, Zhao S, Sangiovanni{-}Vincentelli AL, Keutzer K, Gong B (2019)
  Domain randomization and pyramid consistency: Simulation-to-real
  generalization without accessing target domain data. In: {ICCV}, pp
  2100--2110

\bibitem[{Zhang et~al.(2021{\natexlab{a}})Zhang, Cui, Xu, Zhou, He, and
  Shen}]{DBLP:conf/cvpr/Zhang0XZ0S21}
Zhang X, Cui P, Xu R, Zhou L, He Y, Shen Z (2021{\natexlab{a}}) Deep stable
  learning for out-of-distribution generalization. In: {CVPR}, pp 5372--5382

\bibitem[{Zhang et~al.(2022)Zhang, Xu, Xu, Liu, Cui, Wan, Sun, and
  Li}]{DBLP:journals/corr/abs-2203-14387}
Zhang X, Xu Z, Xu R, Liu J, Cui P, Wan W, Sun C, Li C (2022) Towards domain
  generalization in object detection. CoRR abs/2203.14387

\bibitem[{Zhang et~al.(2021{\natexlab{b}})Zhang, Wang, Wang, Zeng, and
  Liu}]{DBLP:journals/ijcv/ZhangWWZL21}
Zhang Y, Wang C, Wang X, Zeng W, Liu W (2021{\natexlab{b}}) Fairmot: On the
  fairness of detection and re-identification in multiple object tracking. Int
  J Comput Vis pp 3069--3087

\bibitem[{Zhao et~al.(2021)Zhao, Zhong, Yang, Luo, Lin, Li, and
  Sebe}]{DBLP:conf/cvpr/ZhaoZYLLLS21}
Zhao Y, Zhong Z, Yang F, Luo Z, Lin Y, Li S, Sebe N (2021) Learning to
  generalize unseen domains via memory-based multi-source meta-learning for
  person re-identification. In: {CVPR}, pp 6277--6286

\bibitem[{Zhao et~al.(2022)Zhao, Zhong, Zhao, Sebe, and
  Lee}]{DBLP:conf/eccv/ZhaoZZSL22}
Zhao Y, Zhong Z, Zhao N, Sebe N, Lee GH (2022) Style-hallucinated dual
  consistency learning for domain generalized semantic segmentation. In: {ECCV}
  {(28)}, Springer, Lecture Notes in Computer Science, vol 13688, pp 535--552

\bibitem[{Zheng et~al.(2017)Zheng, Zhang, Sun, Chandraker, Yang, and
  Tian}]{DBLP:conf/cvpr/ZhengZSCYT17}
Zheng L, Zhang H, Sun S, Chandraker M, Yang Y, Tian Q (2017) Person
  re-identification in the wild. In: {CVPR}, pp 3346--3355

\bibitem[{Zhong et~al.(2020)Zhong, Wang, and Zhang}]{DBLP:conf/cvpr/ZhongWZ20}
Zhong Y, Wang X, Zhang S (2020) Robust partial matching for person search in
  the wild. In: {CVPR}, pp 6826--6834

\bibitem[{Zhong et~al.(2022)Zhong, Zhao, Lee, and
  Sebe}]{DBLP:conf/nips/ZhongZLS22}
Zhong Z, Zhao Y, Lee GH, Sebe N (2022) Adversarial style augmentation for
  domain generalized urban-scene segmentation. In: NeurIPS

\bibitem[{Zhou et~al.(2022)Zhou, Liu, Qiao, Xiang, and Loy}]{zhou2022domain}
Zhou K, Liu Z, Qiao Y, Xiang T, Loy CC (2022) Domain generalization: A survey.
  IEEE Transactions on Pattern Analysis and Machine Intelligence

\end{thebibliography}
}

\end{document}